\documentclass{article}

\usepackage{listings}
\usepackage{arxiv}
\usepackage{adjustbox}
\usepackage[utf8]{inputenc} 
\usepackage[T1]{fontenc}    
\usepackage{hyperref}       
\usepackage{url}            
\usepackage{booktabs}       
\usepackage{amsfonts}       
\usepackage{nicefrac}       
\usepackage{microtype}      
\usepackage{lipsum}		
\usepackage{graphicx}
\usepackage{natbib}
\usepackage{doi}
\usepackage{bm}
\usepackage{rotating}
\usepackage{tikz}
\usepackage{caption,color,float}

\usepackage[bb=px]{mathalfa} 

\usepackage[varvw]{newtxmath}       
\usepackage{amsmath}
\mathchardef\mhyphen="2D 
\usepackage{booktabs}
\usepackage{multirow}
\usepackage{comment}
\usepackage{diagbox}
\usepackage{soul}
\usepackage{fancyhdr}
\usepackage{wrapfig}
\usepackage{graphicx}

\usepackage{subcaption}

\usepackage{subcaption}

\title{Improved generalization with deep neural operators for engineering systems: Path towards digital twin}

\author{ 	{Kazuma ~Kobayashi} \\
	Nuclear, Plasma \& Radiological Engineering\\
	University of Illinois at Urbana-Champaign\\
	Urnaba, IL 61801, USA \\
 	\And
       {James ~Daniell} \\
	Nuclear Engineering and Radiation Science\\
	Missouri University of Science and Technology\\
	Rolla, MO 65409, USA \\
	\And
      {Syed Bahauddin ~Alam} \\ 
	Nuclear, Plasma \& Radiological Engineering\\
	University of Illinois at Urbana-Champaign\\
	Urnaba, IL 61801, USA \\
 }

\renewcommand{\shorttitle}{\textit{Deep neural operators for engineering systems: Path towards digital twin
}}

\hypersetup{
pdftitle={A template for the arxiv style},
pdfsubject={q-bio.NC, q-bio.QM},
pdfauthor={David S.~Hippocampus, Elias D.~Striatum},
pdfkeywords={First keyword, Second keyword, More},
}

\begin{document}
\maketitle

\begin{abstract}

Neural Operator Networks (ONets) represent a novel advancement in machine learning algorithms, offering a robust and generalizable alternative for approximating partial differential equations (PDEs) solutions. Unlike traditional Neural Networks (NN), which directly approximate functions, ONets specialize in approximating mathematical operators, enhancing their efficacy in addressing complex PDEs.In this work, we evaluate the capabilities of Deep Operator Networks (DeepONets), an ONets implementation using a branch–trunk architecture. Three test cases are studied: a system of ODEs, a general diffusion system, and the convection–diffusion Burgers’ equation. It is demonstrated that DeepONets can accurately learn the solution operators, achieving prediction accuracy (R2) scores above 0.96 for the ODE and diffusion problems over the observed domain while achieving zero-shot (without retraining) capability. More importantly, when evaluated on unseen scenarios (zero-shot feature), the trained models exhibit excellent generalization ability. This underscores ONets’ vital niche for surrogate modeling and digital twin development across physical systems. While convection–diffusion poses a greater challenge, the results confirm the promise of ONets and motivate further enhancements to the DeepONet algorithm. This work represents an important step towards unlocking the potential of digital twins through robust and generalizable surrogates.

\end{abstract}

\section{Introduction}
\label{sec:intro}

Machine Learning (ML) models have become a popular and important tool for solving engineering problems in complex systems. Due to this growth, new ML techniques are being developed to solve problems that other traditional or ML models need to be equipped with. One method receiving significant development is Operator Learning. While traditional ML models such as Neural Networks (NNs) approximate a solution by mapping one function to another, operator learning strategies opt to approximate an operator which can more generally map a relationship with no regard to input or output dimensionality \cite{DeepONet_Lu, operator_maps_kovachki, architectures_algorithms_wang, fourier_pde_li, deeponet_heat_lu, deeponet_stochastic_souvik}. Utilizing this response space conveys benefits in engineering problems that occur over a specific space, specifically physical spaces that can be effectively digitally approximated or complex spaces in which the parameters interact significantly. These models, when developed, are commonly referred to as Operator Networks (ONets). Due to their displayed strengths, ONets are primed to find an impactful and vital niche in computational modeling for engineering solutions.

One of the primary benefits of operator learning is its capability to map functions over a possibly infinite-dimensional space. NNs map inputs directly to outputs, resulting in an undefined domain unless normalized by activation functions \cite{operator_maps_kovachki}. This can result in unreliable predictions when providing inputs outside of bounds used for training \cite{sparse_obs_zhu}. Conversely, by defining the space over which the approximated operator performs, a holistic view of the modeled system or phenomena can be developed \cite{deeponet_heat_lu, bubble_dynamics_lin}. Functioning over a defined domain also simplifies physical engineering problems. 2 or 3-dimensional spaces can be simply handled by the model and used to represent a physical cross-section or volume \cite{deeponet_heat_lu, fem_operator_yin, bubble_growth_lin}. Furthermore, this allows the operator to learn effects from sensor readings both locally and globally \cite{deeponet_heat_lu, fourier_multiphase_wen, fno_eddy_li}. 

However, the unique architecture of operator learning models requires inputs and training to be handled differently from traditional NNs. Domain and feature information must be input and defined separately \cite{DeepONet_Lu, cost_accuracy_hoop}. Consequently, operator learning exacerbates the data requirement problem already present in NNs \cite{deeponet_stochastic_souvik, cost_accuracy_hoop, operator_uncertainty_pickering}. Furthermore, data preprocessing becomes more robust due to the input and output formats. Larger amounts of data result in large computational requirements in training cycles. RAM bottlenecks are common due to the spatial information required in both the input and output since the domain is provided with the observed function. These issues prevent operator learning models from being used universally in problems where standard NNs or ML models may be more efficient.

Due to the unique benefits of operator learning compared to other ML models, these techniques are expected to be used for modeling physical spaces such as cross-sections or volumes. The strengths of operator learning, such as its continuous nature and predefined space, contribute positively to the analysis of physical systems, especially high-importance systems which may be difficult to maintain instrumentation in \cite{fourier_pde_li, deeponet_heat_lu, fem_operator_yin}. Diffusion problems are primed for operator learning since the domain of the problem can be defined by the developer, and a prediction can be obtained at any point inside that domain \cite{deeponet_heat_lu, fourier_operator_kovachki, wavelet_parametric_pde_souvik}. This type of problem is also beneficial since visualization in this space is simple and intuitive. Additionally, diffusion problems can be posed in multiple dimensions, and data can be generated or collected more easily than in other data-driven problems \cite{fourier_pde_li, deeponet_heat_lu}. Materials analysis will likely be an area of ONet learning using 2-dimensional cross sections. Materials problems such as stress distribution, temperature distribution, cracking, or material phase can be represented in this space, allowing a robust collection of input features through the branch network \cite{deeponet_heat_lu}.

This paper aims to understand the feasibility and capability of DeepONets, a type of operator learning method, in solving generalized engineering problems such as a system of ODEs, diffusion-reaction, and convection-diffusion. Therefore, three test cases are shown using DeepONets. These test cases can be used to demonstrate the capability of DeepONets by operator approximation. Each test case utilizes fully-connected NNs for branch and trunk networks to retain mathematical significance by the generalized Universal Approximation Theorem for operators.

\section{Literature on Operator Learning}
\label{sec:literature}

\subsection{Neural Networks}

Based on the architectural structure of the brain, NNs function by accepting inputs and feeding them through a series of layers containing artificial neurons to make a prediction. NNs aim to increase their prediction accuracy by updating neurons with a pre-defined learning rule, typically by using the prediction error of a single or group of samples \cite{nn_backpropagation_yu, deep_learning_overview_schmidhuber}. This results in an NN's ability to continually approach accurate solutions if given sufficient information and trained effectively. Consequently, NNs are valuable tools for function approximation for problems where a function is not known or analytically solvable but where data is available \cite{tanh_approx_anastassiou, nn_approx_cardaliaguet, nn_piecewise_selmic}.

NNs, as one of the most commonly used ML methods for engineering problem solving, are sufficient for simple systems or basic function approximation. NNs function based on the Universal Approximation Theorem, effectively stating that they can capture both linear and nonlinear functions by optimizing the neuron weights to reduce prediction error \cite{nn_approx_cardaliaguet}. This is especially true when modeling a single physical phenomenon or mathematical function.

However, NNs struggle to make accurate predictions when modeling complex problems, such as those with many component phenomena or performing outside the data used to train the model. This can become problematic for simple NNs when training data is not available in the same regions as desired predictions or where the individual phenomena cannot be distinguished from one another \cite{ruthotto2020deep}. While multiple NNs can be used to capture system behavior in different regions or for different phenomena, information about the interaction between these locations can be lost, and the development time for these models increases.

By examining these drawbacks, it can be seen that an ML model that can holistically capture a system with the same self-optimization capabilities of NNs is desirable for complex engineering systems. More specifically, data-driven models that can predict system behavior more generally than the functional level would be useful for engineering problem-solving and analysis.

\subsection{Modern Operator Learning Methods}

Operator learning methods have begun to be explored for scientific applications with three primary types of models: Fourier Neural Operators (FNO) \cite{operator_maps_kovachki,fourier_pde_li,fno_acoustic_guan}, Wavelet Neural Operators (WNO) \cite{wavelet_parametric_pde_souvik, multifidelity_wavelet_souvik}, and DeepONets \cite{DeepONet_Lu, architectures_algorithms_wang, pino_goswami}. Each model satisfies the Universal Approximation Theorem for Operators, although they are slightly different from a design perspective\cite{DeepONet_Lu, operator_comparison_lu}. This allows the models to achieve the goal of producing an approximation that is continuous in an infinite dimensional output space\cite{Gorban2002, Chen1995, cost_accuracy_hoop, operator_uncertainty_pickering}.

FNOs and WNOs utilize a different approach from DeepONets, which utilize concepts native to deep learning NNs to produce an approximation \cite{operator_maps_kovachki, fourier_operator_kovachki, operator_comparison_lu}. FNOs and WNOs use transformations to produce an approximation in the infinite-dimensional response space \cite{operator_maps_kovachki, fourier_operator_kovachki, operator_comparison_lu}. FNOs and WNOs both use a NN based front-end for filtering information and generalizing inputs, which are then transformed into the response space using a Fourier transform for FNOs or a generalized wavelet transform for WNOs \cite{fourier_pde_li, operator_maps_kovachki}. The NN portions of the architecture for these models are trained using backpropagation methods in order to approximate the desired operator more accurately. These models are useful for multiphase flow and materials deformation problems, as well as general PDE solutions in which information about the phenomenon is known \cite{fourier_pde_li, fourier_multiphase_wen, fno_implicit_materials_you, fno_eddy_li, wavelet_parametric_pde_souvik}. Furthermore, due to the ability to temporally synchronize information from data, WNOs are useful for uncertainty quantification over time \cite{operator_uncertainty_pickering, multifidelity_wavelet_souvik, wavelet_parametric_pde_souvik}.

DeepONets directly utilize methods native to deep learning problem solving instead of the transformations in FNOs and WNOs and inherit directly from NN design \cite{DeepONet_Lu, onet_elliptic_eq_marcati}. A dot product combines information from the branch and trunk networks in the model architecture to approximate the operator \cite{architectures_algorithms_wang, fem_operator_yin}. This methodology allows for mapping finite inputs to the infinite response space \cite{DeepONet_Lu, architectures_algorithms_wang, operator_comparison_lu}. Additionally, DeepONets have seen active development for some scientific applications through Python libraries such as DeepXDE \cite{lu2021deepxde}. This has allowed scientists and engineers to develop ML-based techniques for general PDE approximation \cite{onet_elliptic_eq_marcati, bubble_dynamics_lin, extreme_forecasting_pickering}.

Furthermore, the DeepONet architecture has been modified for additional applications due to its relative simplicity. Similar to the Physics Informed Neural Network (PINN), Physics Informed Neural Operators (PINOs) \cite{pino_goswami, pde_pino_li} have been developed to improve physics synchronization for applications in which some physics knowledge is known \cite{pino_goswami}. Since DeepONet approximations can be examined continuously, the data-physics fusion approach can be used to develop highly reliable models of systems \cite{deeponet_heat_lu}. Furthermore, for applications with minimal data, multi-fidelity approaches to operator learning using DeepONets have been developed \cite{deeponet_heat_lu}. These methods utilize low-fidelity data from equations or empirical models alongside sensor data to form an approximation based on both datasets \cite{multifidelity_wavelet_souvik, deeponet_heat_lu}.

\subsection{Research Gaps}

The primary research gap with modern operator learning methods is due to the types of systems and associated applications. Due to their recent development, operator learning methods focus on analytically solvable problems for which data can be generated. This is because operator learning methods can be applied without experimental data and easily checked against ground truth values with these problems. However, data-driven prediction for unknown or unsolved systems has not been explored using operator learning principles. While this is the predicted use case of approximated operators from these models, the large data requirement for producing accurate and robust results poses a challenge in most industries.
Additionally, available sensors and instrumentation in engineering systems result in a data-sparse environment where sensor locations remain constant. Therefore, robust operator approximation methods must be developed for data-driven methods in data-sparse environments that can safely make predictions, ideally without being forced to make assumptions. This is where the DeepONet, a natively data-driven method, can be utilized.

\section{Operator Learning \& Digital Twin Applications}

As modern engineering systems are developed, so do approaches to explaining these systems. Advances in instrumentation and controls have led to systems that can collect significant amounts of information compared to their previous counterparts. This data and information have been used for validation and experimental correlations. However, the rapid emergence of data-driven explanation methods, including ML models, has expanded the applications and value of the data. Since information is becoming an increasingly vital part of engineering systems, methods to aggregate and adequately explain this information are necessary \cite{bonney2022development}. This is the role of digital twins (DTs) in modern engineering systems.

By combining previous knowledge, data-driven methods, and real-time information, a DT may be developed to encapsulate knowledge about a system and present it in a way humans can understand. Additionally, with physics knowledge and data-driven methods, advanced DTs can utilize information to make predictions about the system or produce information about system components that sensors cannot observe. However, suppose information from a DT is required during operation. In that case, it must be developed such that computation time is sufficiently short enough to produce these results when the information is needed \cite{kabir2010non,kabir2010theory,kabir2010watermarking,kabir2010loss}. This can be challenging due to the accuracy tradeoff for multi-physics simulations, which typically require increased computation time for high-accuracy results. Modern ML methods can help solve this issue by providing computationally cheap estimates for system parameters, which can be used alongside physics equations to provide sufficiently accurate information for the DT \cite{kobayashi2022surrogate,kobayashi2022data}.

Ultimately, DTs in complex engineering systems are plausible because of the increased utility of information. More specifically, the Internet of Things (IoT) approach to modern systems increases the viability and capability of DTs in engineering systems. A framework for producing DT-type depictions of complex systems can begin to form by linking sensors and aggregating this information. IoT approaches network information, sensors, and system components in a way that can be used externally. DTs can be used alongside the IoT to synchronize information, which is used alongside the approaches above to predict, extrapolate, and generalize data to produce a more complete picture of the system \cite{bonney2022development}.

By their design criteria, DeepONets lend themselves to DT development in complex engineering systems. Since DeepONets can map infinite-dimensional function spaces to infinite-dimensional output spaces, the development of supporting DT modules can be generalized for a system, and additional sensors or components can be added without overhauling model development. Furthermore, DeepONets can robustly predict system behavior with a well-defined system space due to their ability to generalize phenomena versus direct equation mapping from traditional NNs or other finite-dimensional ML models \cite{bubble_dynamics_lin}. DeepONets could also be used as supplementary controller modules inside of DTs, which aggregate information from single-physics models and determine system behavior and interaction without using computationally expensive multi-physics modeling directly \cite{kobayashi2022surrogate,kobayashi2022data}. In this sense, developing DTs for complex systems is directly correlated to developing effective and computationally cheap neural operators for advanced systems.

\begin{figure}[!htbp]
    \centering
    \includegraphics[width=17cm, angle=90]{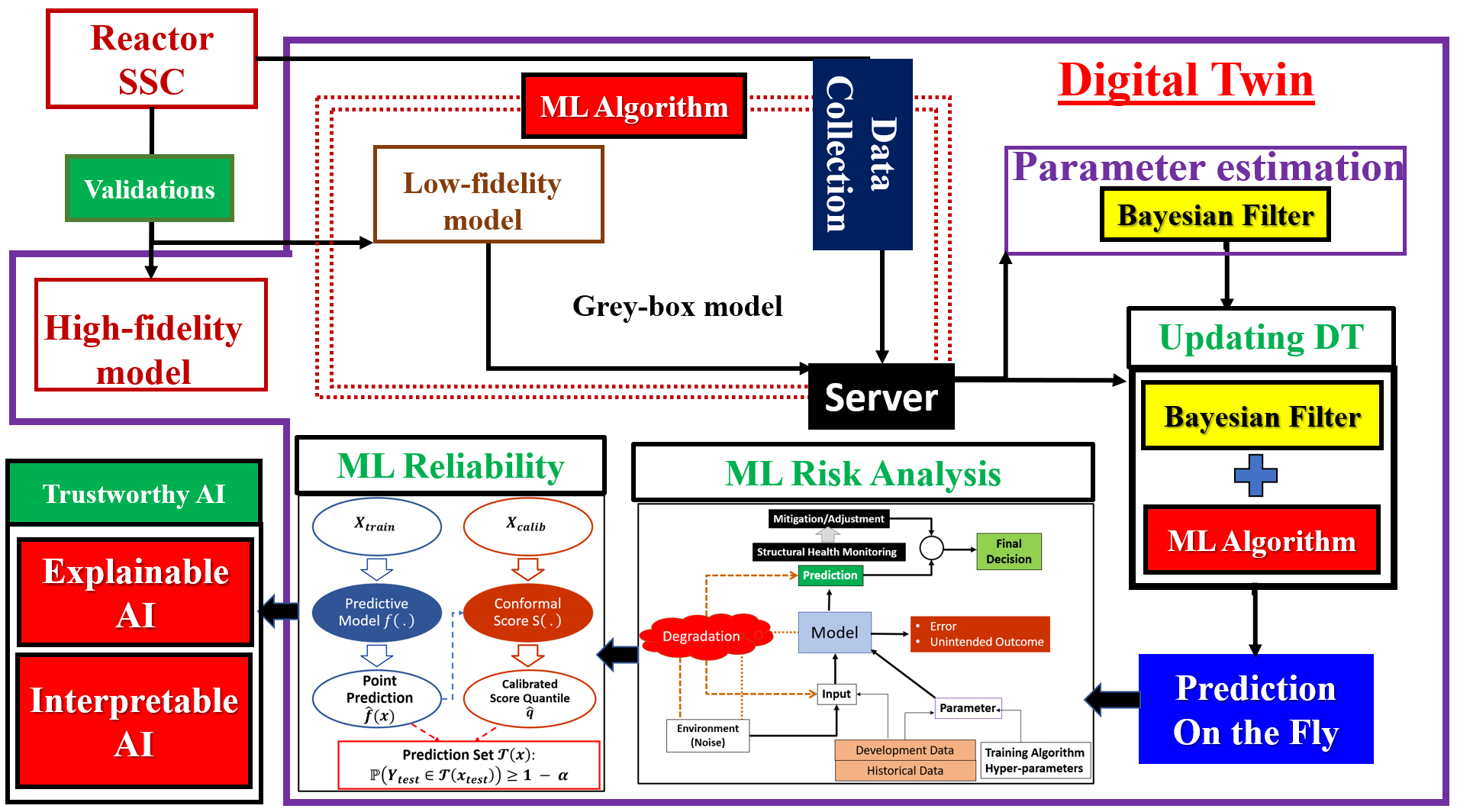}
    \caption{Intelligent Digital Twin Framework with Explainable AI and Interpretable ML module. The diagram shows the ML components (Red Boxes) exploited in different segments of the digital twin framework \cite{kazuma_eaai}.}
       \label{fig:DTpic3}
\end{figure}

There are five essential components in our proposed DT framework \cite{kobayashi2023explainable}: (1) prediction module, (2) Real-time update module for ``On the Fly" temporal synchronization,  (3) data processing module, (4) visualization module, and (5) decision-making module. Also, there are two important components to ensure trustworthiness and reliability, which are (6) ML Risk analysis and (7) ML Reliability, and (8) Trustworthy AI. Figure \ref{fig:DTpic3} shows the proposed Intelligent DT Framework \cite{kobayashi2023explainable}, and initially, this methodology has been proposed by \cite{garg2022physics}. According to the figure, the ML algorithm collects data from a physical asset and a high-fidelity simulation that integrates both high- and low-fidelity data streams. The ML algorithm analyzes the multi-fidelity data to make predictions of interest, with the predicted output stored on a server. The prediction and system update modules require sophisticated ML algorithms. A key component is the system update module based on temporal synchronization, which utilizes a Bayesian filter combined with an ML algorithm to enable real-time, on-the-fly temporal synchronization \cite{kobayashi2023explainable}. Figure \ref{fig:DTpic3} shows the ML components (Red Boxes) exploited in different segments of the DT framework for prognostic. Overall, the system leverages Bayesian statistical modeling fused with ML techniques for prediction and continuously updating the model state based on streaming multi-fidelity data from the physical system and simulations. 

In addition, Figure \ref{fig:DTpic3} shows an explainable and trustworthy AI component. To meet NIST's \cite{tabassi2023artificial} definition requirement (released on 03/17/2022) on “Trustworthy AI \& ML Risk Management Framework,” \cite{pimlModelInterpret1Part1, pimlModeldiagnostic2Part2} specific tests need to be performed such as (1) Robustness Test under covariate perturbation and worst sampler resilience to measure model sensitivity to variations in uncontrollable factors. (2) Prediction Reliability Test by split conformal prediction, segmented bandwidth, and distribution shift (Reliable vs. unreliable) to ensure the model consistently generates the same results within the bounds of acceptable statistical error, and (3) Resilience Test for prediction performance degradation by worst-case subsampling and under out-of-distribution scenarios. Furthermore, assessing the reliability  \cite{pimlModelInterpret1Part1, pimlModeldiagnostic2Part2} of machine learning predictions is important for understanding where a model makes less reliable forecasts. Wider prediction intervals indicate less reliable predictions. Quantifying prediction reliability can be done with split conformal prediction, assuming exchangeability. For binary classifiers, reliability diagrams show probability calibration relative to empirically observed success rates. Validation trustworthiness is confirmed by incrementally adding additional noise to all measured parameters to determine if the signal-to-noise (SNR) ratios were to degrade by an additional X\% that will assess the comfort margin. Furthermore, to ensure trustworthiness, it requires managing both aleatory and epistemic uncertainties while explicitly accounting for the multiscale uncertainty in specific problem cases. 

The proposed digital twin also aims to implement explainable AI (XAI) for decision-making and observation of anomalies by adversarial robustness and semantic saliency, currently under development by the authors. These methods, if successful, will assess the theoretical basis and practical limits of the explainability of the neural operator algorithm used. In our future work, we will employ a proof-based method to probe all hidden layers of the neural operator model, identifying the most important layers and neural networks involved in predictions to ensure prediction explainability.

Although the components of DT vary depending on its purpose, the standard functions that form its core are prediction and system parameter updating using real-time data sent from sensors installed in the physical system. Because of its property of mapping between input and output functions, ONets can be a viable option in implementing these essential functions. In this section, we will briefly explain prediction, system parameter update, and utilization of ONets to them.

One of the purposes of DT is to make a real-time prediction of the system's behaviors using sensor data from a physical system. Although DT is sometimes confused with simulation, this difference distinguishes the two. In a simulation, the user provides input data to the simulation solver (e.g., FEM, FDM, FVM), such as the operating conditions of the system and material properties, and obtains predictions as output. However, the computational cost of simulation is a critical issue when targeting complex systems. Simulating the fluid velocity profile in a complex system using a FEM solver can take hours. Therefore, even if the solver solves the system state using data from the sensors, there is a lag until the calculation results are obtained; DT aims to minimize this lag as much as possible and predict the system state immediately. To solve this issue, surrogate modeling methods have been utilized as a new approach in recent years \cite{kobayashi2022surrogate, daniell2022physics, rahman2022leveraging}.

\begin{figure}[!htbp]
    \centering
    \includegraphics[scale=0.5]{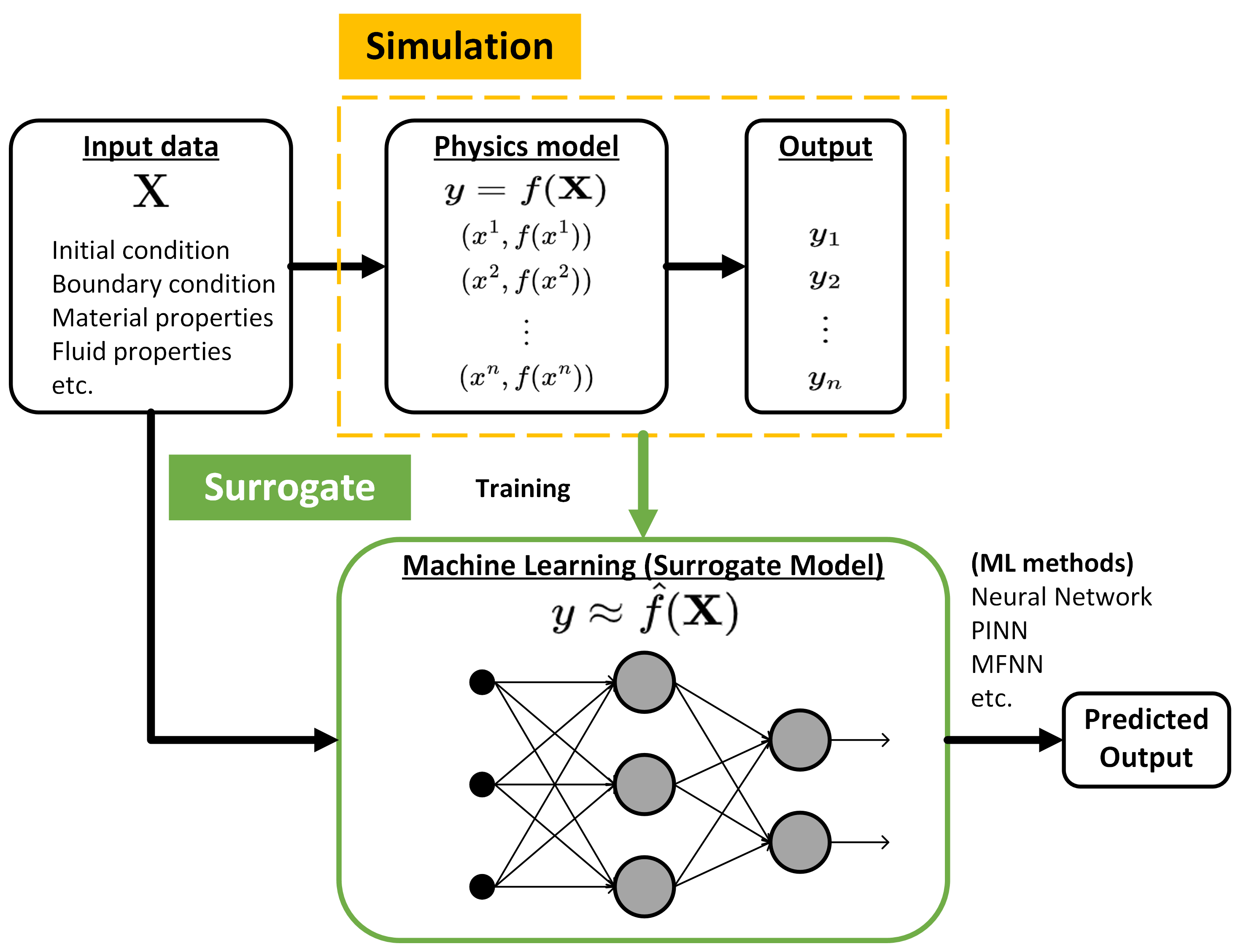}
    \caption{Concept of surrogate modeling method. The surrogate model can return predictions immediately when the input variables are given. The demand for conventional simulations has not disappeared to prepare the training data for ML modeling.}
    \label{fig:surrogate}
\end{figure}

In this approach, a solver predicts the system state under various input variable conditions (initial conditions, boundary conditions, material properties, etc.) in advance and builds an ML model using these as training data. Figure \ref{fig:surrogate} shows the relationship between conventional simulation and surrogate modeling. The constructed ML model returns immediate predictions for the input variables. Therefore, it can satisfy DT's requirements. Although there are several supervised ML methods (NN, PINN, MFNN), ONets could be a new option when building this surrogate model.

\begin{figure}[!htbp]
    \centering
    \includegraphics[width=13cm]{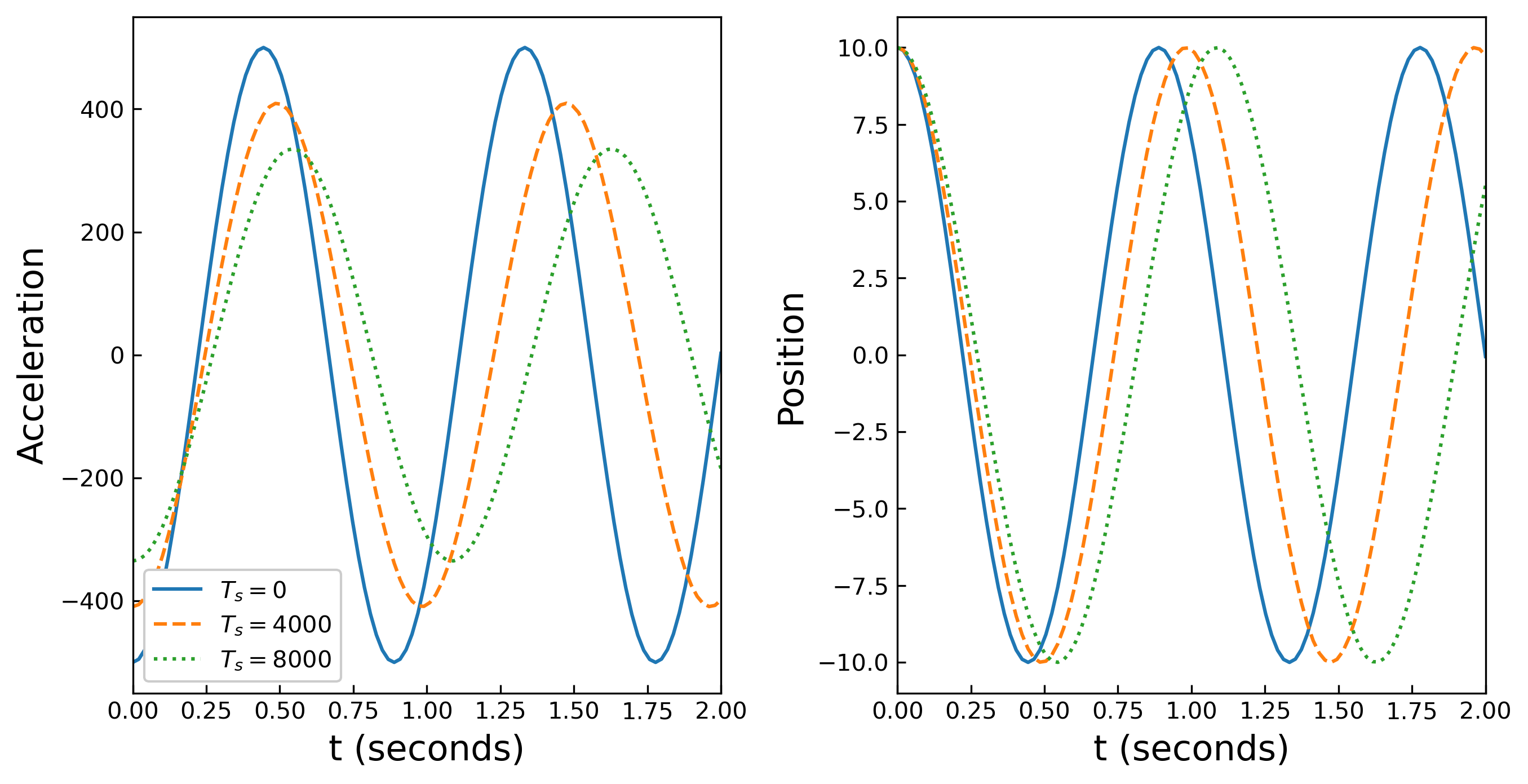}
    \caption{Acceleration and position of the mass-spring system described by equation \ref{eq:mass-spring}. $T_{s}=0$ represents the ideal solutions. $T_{s} = 4000$ and $T_{s}=8000$ represent system operation time in days. }
    \label{fig:update-demo}
\end{figure}

DT requires temporal synchronization of system parameters between the digital and physical systems to predict future system states. For all practical purposes, the system's lifecycle is much slower (in months or years) than the system dynamics' time scale. Therefore, a temporal synchronization of system variables in a slow time scale is necessary for prediction over the operational life of a system. The importance of the system parameter update is explained in the following. Let's assume the system is a mass-spring system and described as a function of dynamics' time-scale ($t$) and slow time-scale ($t_{s}$):

\begin{equation}
    M(t_{s}) \frac{\partial ^{2}X(t, t_{s})}{\partial t^{2}} + K(t_{s})X(t,t_{s}) = 0
\label{eq:mass-spring}
\end{equation}
\noindent
where $M(t_{s})$ is the mass and is assumed to be a constant over the lifecycle, $K(t_{s})$ is the spring constant, and $X(t,t_{s})$ is the system state. If the value of the spring constant is given, the system state can be obtained by solving ODE. Figure \ref{fig:update-demo} shows the sample solutions as solid blue lines. However, assuming this system will operate for approximately 20 years, system variables may change gradually on a slow time scale. Various environmental factors, such as temperature, pressure, and even irradiation, can cause the degradation of system parameters. The orange and green solid lines in Figure \ref{fig:update-demo} represent the system states when the decreasing of the spring constant is taken into account. As Figure \ref{fig:update-demo} shows, system parameters over long time scales affect the system state and require their updating to compensate.

\begin{figure}[!htbp]
    \centering
    \includegraphics[width=13cm]{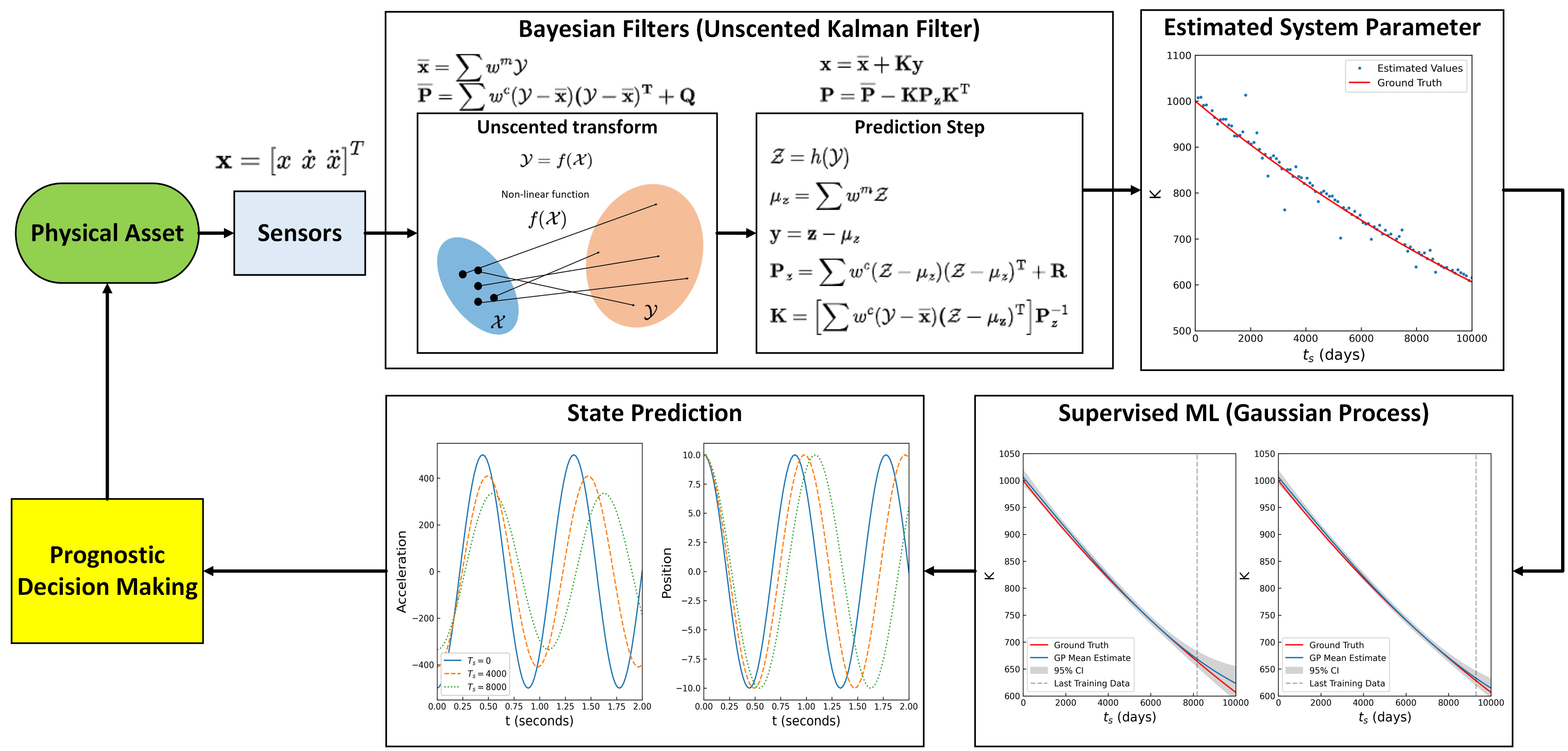}
    \caption{Concept of update module in DT - synchronizing the physical system and DT \cite{kobayashi2023explainable}.}
    \label{fig:update}
\end{figure}

The use of Bayesian filters is an option to implement system parameter updates. The Unscented Kalman Filter (UKF) is expected among the filters due to its superior performance for higher-order nonlinear systems. While the UKF is generally not suited for long operation lifetimes, a new algorithm that employs an analytical resampling process combined with ML is developed \cite{garg2022physics}. Although the details of filter design in this study will not be examined, it is essential to note that the ML method can be used to extend the conventional UKF for DT. While \cite{garg2022physics} employed the Gaussian process (GP) as a supervised ML, they also suggested replacing it with other ML algorithms. The basic concept of system parameter updates is presented in Figure \ref{fig:update}, and we expect to utilize ONets instead of GP.

\section{Governing Principles of Operator Learning}

\begin{figure}
    \centering
    \includegraphics[width=\textwidth]{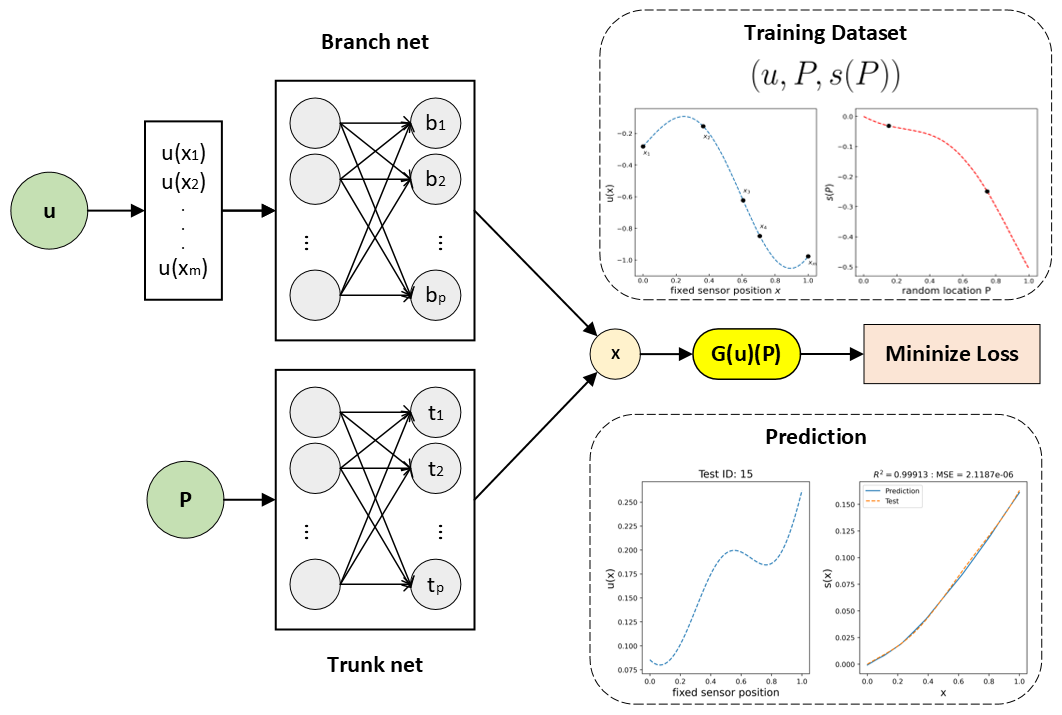}
    \caption{DeepONet Branch-Trunk Architecture following the proposed approach from \cite{DeepONet_Lu}. The training dataset is composed of (1) input function $u(x_{m})$, (2) sampling positions $P$ for the system output, and (3) system output $s$ at the position $P$.}
    \label{fig:ONet_Arch}
\end{figure}

Much like NNs, DeepONets were developed using the concept of universal
approximation, although in this case, the Universal Approximation Theorem for
Operators is examined~\cite{DeepONet_Lu}. NNs have been traditionally utilized to map inputs into a designated function space. In contrast, DeepONets are designed to transform information from functional forms into operators applicable within an arbitrary domain. Within the DeepONet architecture, input functions undergo discretization via sampling at specific locations, represented as $\{x_{1}, x_{2}, \dots, x_{m}\}$, where $m$ denotes the total number of discretized points. This approach enables DeepONet to adeptly manage two types of network inputs: $[u(x_1), u(x_2), \dots, u(x_m)]^{T}$ and $P$, which respectively correspond to the sampling positions and system output $s$. The efficacy of this method is grounded in the Universal Approximation Theorem for Operators, formalized as Eq.~\ref{UAT}.

\begin{equation}
\label{UAT}
    \left| G(u)(P) - \sum_{k=1}^{l}\underbrace{\sum_{i=1}^{n}c_{i}^{k}\sigma \left( \sum_{j=1}^{m}\xi_{ij}^{k}u(x_{j}) + \theta_{i}^{k} \right)}_{\text{branch}} \underbrace{\sigma (w_{k}\cdot P + \zeta_{k})}_{\text{trunk}}   \right| < \epsilon
\end{equation}
This equation shows that a prediction can be generated by utilizing neurons for information filtering, which is then compared to the ground truth value, much like NNs. Unlike NN training, an operator $G(u)(P)$ is tested against instead of a sensor value. In this case, $G(u)(P)$ represents the operator $G$ performing on sensor value(s) $u$ over domain $P$. The summation used in \eqref{UAT} is similar to the standard Universal Approximation Theorem, which implies that a function object like a NN could be used in this location. Additionally, since $u$ in this equation represents some function, the feature space can be infinitely large.

\begin{equation}
\label{balanced UAT}
G(u)(P) \approx \sum_{k=1}^{l}\underbrace{b_{k}\left( u(x_{1}), u(x_{2}), ..., u(x_{m}) \right)}_{\text{branch}}\underbrace{t_{k}(P)}_{\text{trunk}}
\end{equation}

\begin{equation}
\label{UAT for operator}
    \left| G(u)(P) - \langle \underbrace{\bm{g}\left( u(x_{1}, u_{2}, \cdots, u(x_{m}) \right)}_{\text{branch}}, \underbrace{\bm{f}(P)}_{\text{trunk}} \rangle  \right| < \epsilon
\end{equation}
The Universal Approximation Theorem for Operators can be manipulated such that the sensor and domain information can be separated to draw implications about model architectures required for predictions. Represented as \ref{balanced UAT}, the sensors and domain can be input separately to a model \cite{DeepONet_Lu,Chen1995}. For DeepONets, a branch-trunk architecture is used. The branch is applied to encode sensor information while the trunk is applied to encode domain information \cite{DeepONet_Lu, Chen1995}. By utilizing this architecture, the operator $G$ can be approximated.
Furthermore, by utilizing \ref{UAT for operator}, predictions from the operator approximation can be compared to ground truth observations to develop a way to train the model. A single NN is used for the domain information for the trunk $\mathbf{f}$, while one or multiple NNs are used to encode sensor information from the function for the branch $\mathbf{g}$. This ‘encoding’ is handled internally by the DeepONet by filtering the input information through the section of the model aimed at using that data. Two different branch architecture types can be used for DeepONet development; ‘stacked’ DeepONets utilize multiple NNs in the branch while ‘unstacked’ DeepONets utilize only a single NN for the branch \cite{DeepONet_Lu, operator_comparison_lu}. The branch and trunk networks are then utilized to approximate the operator for the system, which is trained using the loss associated with the prediction. This architecture can be seen in figure \ref{fig:ONet_Arch}. In this case, $u$ represents the observed function, and $P$ represents the domain over which it occurs. $G(u)(P)$ represents the prediction generated via the operation of the approximated operator ($G$) on the received data, $u$ and $P$. This is iterated over the training process with exposure to multiple samples (combinations of $u$ and $P$) to adjust neuron values in the model based on results from the loss function to approach a more accurate operator approximation.

DeepONet input structure was developed to allow any number of observations from $u$ to be utilized. Each sensor on $u$ collects information from the system at those points, which can be used for operator learning to improve approximations. Additionally, input and output dimensions become trivialized by using the concepts from the Universal Approximation Theorem for Operators. The relationship between an infinite dimensional input space and an infinite dimensional output space can be characterized by approximating a generalized operator. This is contrary to traditional ML models, which can be viewed as a parametric map of some finite input function to the output function.

\section{Test Problems \& Results}
In this section, the versatility and effectiveness of DeepONets are demonstrated by applying them to various simple systems. Specifically, three scenarios were investigated: (1) a system of ODEs, (2) a 1-dimensional diffusion-reaction process, and (3) a 1-dimensional convection-diffusion-reaction phenomenon. These case studies showcased the robustness and applicability of DeepONets in various dynamic systems, highlighting their potential as a powerful modeling tool.

For each problem, the solutions obtained from solving the equations with the solver were utilized as training data to construct a surrogate model using DeepONet. The performance of the surrogate model was assessed by calculating several evaluation metrics using the test data. These metrics, including $\rm R^2$ (Coefficient of Determination), MSE (Mean Squared Error), RMSE (Root Mean Squared Error), MAE (Mean Absolute Error), and RMSE/MAE \footnote[1]{When only random noise remains as the error in a model, causing it to conform to a normal distribution, the ratio of RMSE to MAE is approximately 1.253.} provided valuable insights into the accuracy and reliability of the DeepONets surrogate model.

Furthermore, a comparative analysis was conducted between DeepONets and conventional machine learning modeling methods, namely FCN (fully connected neural network) and CNN (convolutional neural network), for test cases with the highest and lowest $\rm R^2$ values. This Comparison allowed for an evaluation of the performance of DeepONets in relation to established modeling approaches.

This comprehensive evaluation showcases the effectiveness of DeepONets as a versatile and accurate modeling approach capable of outperforming traditional machine learning methods in capturing the complexities of dynamic systems. The implementation of DeepONets was done using the scientific machine learning library DeepXDE \cite{lu2021deepxde}.

\subsection{Setup of a System of ODEs}
\label{sec:ode}
In the foundational problem addressed by this study, we scrutinize a system of ODEs encapsulated by equation \ref{eq:ode}. The focal point of this analysis is to discern the operator that governs the relationship between the input function \( u(x) \) and the output state function \( s(x) \) over the domain \( x \in (0,1] \), while honoring the initial condition \( s(0) = 0 \).

\begin{eqnarray}
  \begin{cases}
    &\frac{ds(x)}{dx} = u(x), \,\,\,\,\, x \in (0,1] \\
    &s(0)=0
  \end{cases}
  \label{eq:ode}
\end{eqnarray}

The state function \( s(x) \), arising as a solution to the ODE, is a scalar field with its dimensionality explicitly defined as one. Mathematically, this is characterized by the mapping \( s: \mathbb{R} \to \mathbb{R} \), signifying that for each real-valued point \( x \) within our domain, there is a corresponding real-valued scalar state \( s(x) \).

\begin{figure}[!htbp]
    \centering
    \includegraphics[width=13cm]{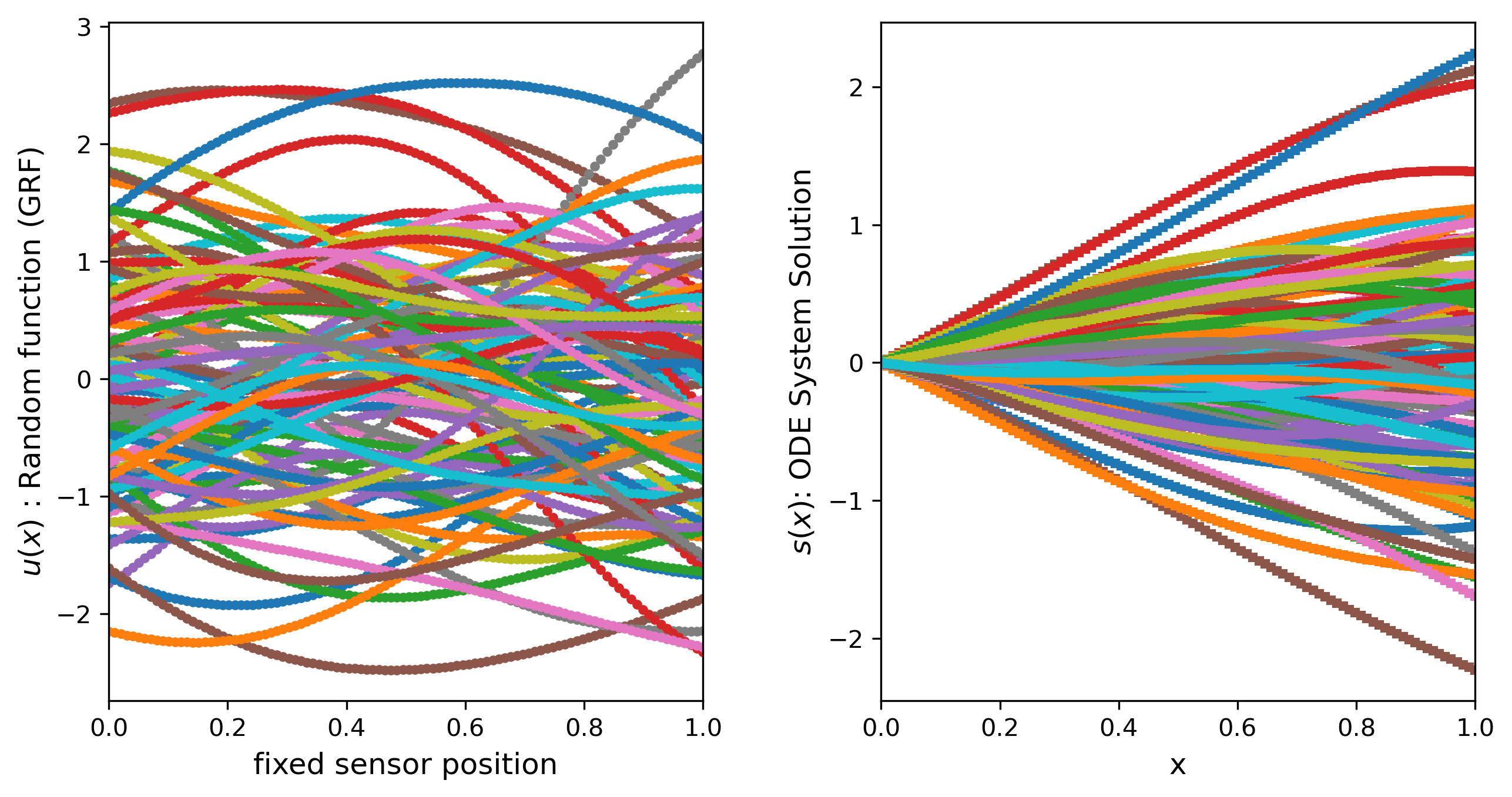}
    \caption{Test Case from \ref{sec:ode}: Left pane represents 150 input function $u(x)$ sampled from 1-dimensional GRF. The numerical solution obtained using the RK method is represented in the right pane.}
    \label{fig:ode_train}
\end{figure}

To curate the training data for the branch network, we opted for a 1-dimensional Gaussian random field (GRF) as the stochastic input function \( u(x) \). A compendium of 150 samples of \( u(x) \) was generated, each uniformly probed at predefined sensor nodes \( \{x_{1}, x_{2}, \cdots, x_{100}\} \), as demonstrated in the left panel of Figure \ref{fig:ode_train}. This systematic sampling approach was devised to secure a consistent and orderly dataset, which is essential for the efficacious training of our neural network.

The training data for the trunk network was generated by solving the ODE using the Runge-Kutta (RK) method with 1,000 steps for each corresponding input function, as depicted in the right panel of Figure \ref{fig:ode_train}. From the obtained solutions, 100 $s(x)$ values were randomly selected for each input function. This selection process aims to capture diverse solution patterns and facilitate comprehensive learning of the operator mapping from input to output functions.

It is worth emphasizing that the 100 random solution sampling locations for each input function were fixed and maintained throughout the dataset. This approach, commonly called "aligned data" \cite{lu2021deepxde}, ensures consistent and comparable training and evaluation of the models.

Similarly, a test dataset was prepared consisting of 1,000 random input functions. This independent test dataset evaluates the model's generalization ability and thoroughly assesses its performance on unseen data.

The architecture of DeepONet is defined as follows: both the branch and trunk networks are fully connected neural networks. The branch network has a size of [100, 40, 40], while the trunk network has a size of [1, 40, 40]. The activation functions utilized in both networks are Rectified Linear Units (ReLU). Weight initialization is performed using the Glorot initialization method, which ensures effective initialization of the network parameters. It is important to note that the same activation functions and initialization methods are employed in the subsequent problems discussed in Sections \ref{sec:dr} and \ref{sec:burger}.

The Adam optimization algorithm is chosen as the optimization method during the training process. The mean $L^{2}$ relative error is used as the evaluation metric to measure the performance of the model. The model is trained for 10,000 iterations. The learning rate for training is set to 0.001, which controls the step size during gradient descent and affects the convergence speed and accuracy of the training process. These choices of optimization algorithm, evaluation metric, and learning rate are consistent across the problems discussed in this paper, ensuring a fair and comparable evaluation of the models. 

When applying traditional NN modeling approaches like FCNs or CNNs to ODEs and partial differential equations (PDEs), separate models need to be built for each specific condition, such as initial and boundary conditions. Consequently, for a problem involving 150 conditions (input function patterns), it would be necessary to construct 150 individual models.

In contrast, ONets, as mentioned earlier, are designed to establish a mapping between input and output functions. As a result, a single DeepONet model can effectively handle multiple conditions within the trained input function domain. This advantage significantly reduces the complexity and computational burden associated with constructing separate models for each condition, making DeepONets a more efficient and versatile modeling approach.

\subsubsection{Results of Ordinary Differential Equation}
The performance evaluation of the DeepONet model on the test data is summarized in Table \ref{tab:ode_all}, providing key summary metrics such as the mean, standard deviation, minimum, and maximum values. The model consistently achieved convergence in terms of $\rm R^2$, with an average value of approximately 0.997, indicating a high level of agreement between the predicted and actual values. This demonstrates the effectiveness of the DeepONet model in capturing the underlying patterns in the data.

\begin{figure}[!htbp]
    \centering
    \includegraphics[width=13cm]{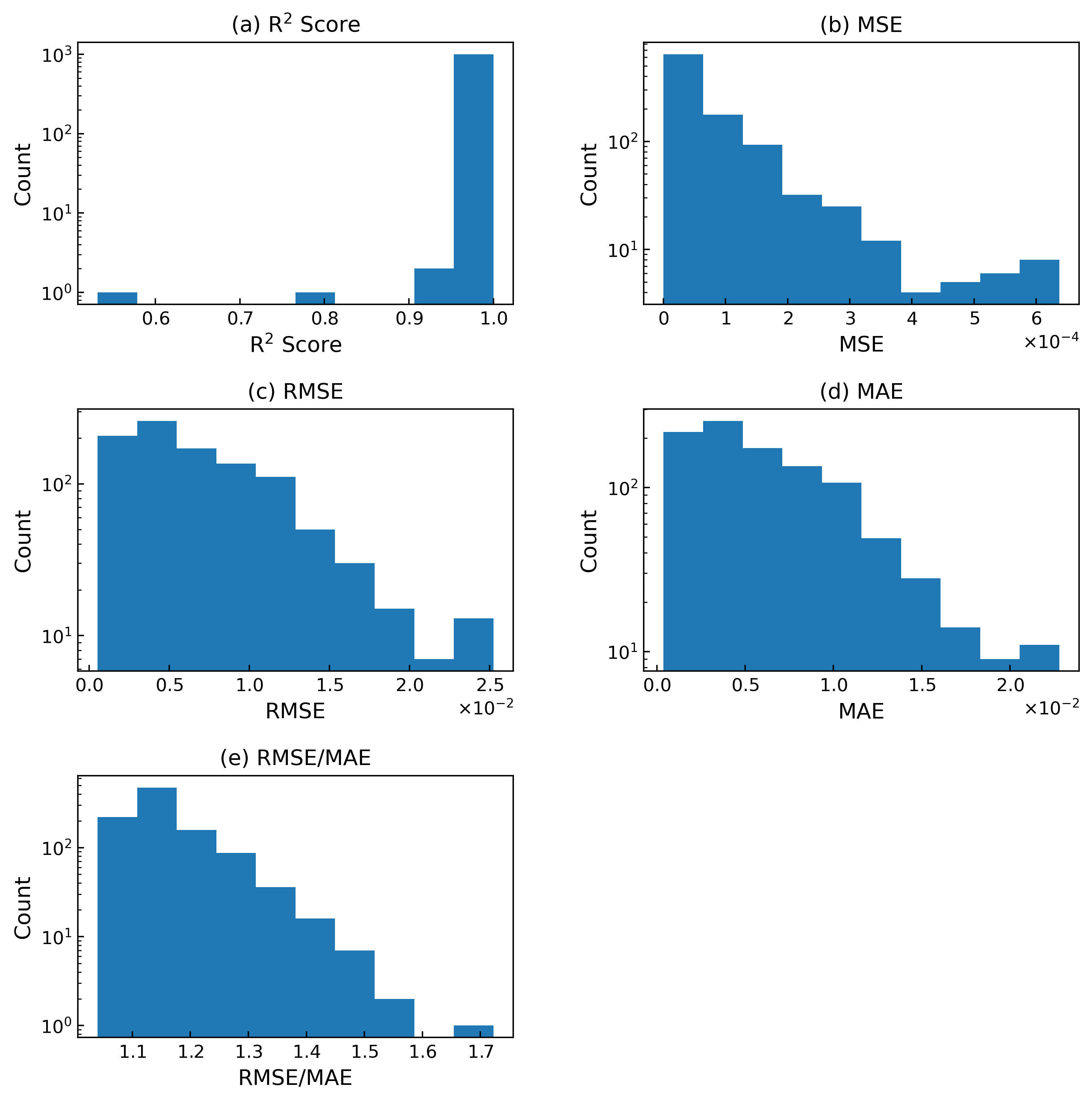}
    \caption{Distribution of each metric over the test
data for the ODE system (Section \ref{sec:ode}).}
    \label{fig:ode_eval_all}
\end{figure}

A detailed analysis of the model's performance is illustrated in Figure \ref{fig:ode_eval_all}, where the Distribution of evaluation metrics is presented. Notably, the $\rm R^2$ scores exhibit strong performance across the test dataset, with the lowest observed value of 0.532. While this suggests that there are certain instances with relatively larger deviations between the predicted and ground truth values, the overall performance of the model remains highly reliable.

Further examination of the metrics reveals that MSE, RMSE, and MAE values are consistently low, ranging from $10^{-4}$ to $10^{-2}$. These results, depicted in Figures \ref{fig:ode_eval_all} (b), (c), and (d), indicate the model's ability to capture the underlying dynamics of the problem accurately. The relatively small magnitudes of these errors highlight the effectiveness of the DeepONet model in accurately predicting the target variable.

The mean ratio of RMSE to MAE, calculated as 1.164 and listed in Table \ref{tab:ode_all}, suggests that there may be a uniform distribution error present in each sample. This information can guide further analysis and improvement of the model to address potential sources of error.

\begin{figure}[!htbp]
    \centering
    \includegraphics[width=13cm]{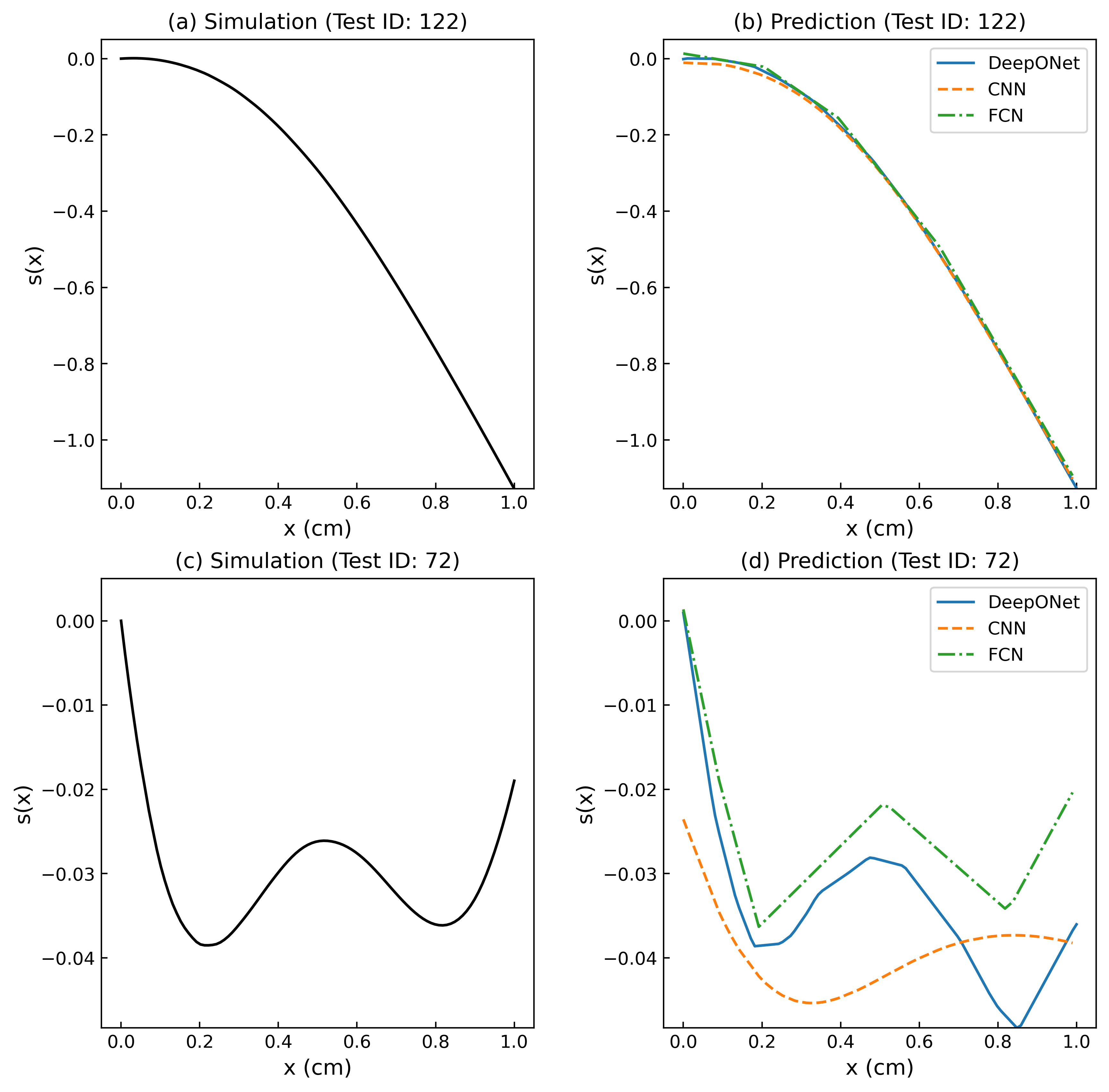}
    \caption{Comparison between simulations and machine learning (ML) predictions. The test cases with IDs 122 and 72 correspond to the highest and lowest $\rm R^{2}$ scores achieved by DeepONet, respectively. The left figures display the simulation results, while the right figures illustrate the predictions made by DeepONet (blue lines), FCN (green lines), and CNN (orange lines). }
    \label{fig:ode_prediction}
\end{figure}

To evaluate the models' performance more comprehensively, a comparison was conducted to determine the highest and lowest prediction accuracies relative to FCN and CNN. These results are summarized in Table \ref{tab:ode_best_worst}, offering valuable insights into the differential performance of DeepONet compared to traditional methods (FCN and CNN). Furthermore, Figure \ref{fig:ode_prediction} presents a specific test case, showcasing the Comparison between simulations and ML predictions. The left figures display the simulation results, while the right figures illustrate the predictions made by DeepONet, FCN, and CNN. Notably, for Test ID 122, which exhibits the highest $\rm R^2$ score, Figures \ref{fig:ode_prediction} (a) and (b) clearly demonstrate that the predictions of DeepONet, FCN, and CNN are in approximate agreement and accurately reproduce the simulation results. However, in the case of Test ID 72, DeepONet exhibits a lower $\rm R^2$ score compared to FCN, and CNN fails to reproduce the simulation result altogether, as shown in Figures \ref{fig:ode_prediction} (c) and (d). These comparative results provide valuable insights into the differential performance of DeepONet and traditional methods for specific test cases.

\begin{table}[!htbp]
\centering
\caption{Overall performance metrics of the DeepONet model for the system of ODEs}
\label{tab:ode_all}
\begin{adjustbox}{width=\textwidth}
\begin{tabular}{@{}llllll@{}}
\toprule
Statistics & $\rm R^2$        & MSE       & RMSE      & MAE       & RMSE/MAE  \\ \midrule
Mean       & $9.974\times 10^{-1}$ & $7.420\times 10^{-5}$ & $7.163\times 10^{-3}$ & $6.271\times 10^{-3}$ & $1.164$ \\
Std        & $1.649\times 10^{-2}$ & $1.009\times 10^{-5}$ & $4.788\times 10^{-3}$ & $4.307\times 10^{-3}$ & $8.022\times 10^{-2}$ \\
Min        & $5.316\times 10^{-1}$ & $2.861\times 10^{-7}$ & $5.350\times 10^{-4}$ & $3.810\times 10^{-4}$ & $1.040$ \\
Max        & $1.000$ & $6.372\times 10^{-4}$ & $2.524\times 10^{-2}$ & $2.279\times 10^{-2}$ & $1.722$ \\ \bottomrule
\end{tabular}
\end{adjustbox}
\end{table}

\begin{table}[!htbp]
\centering
\caption{Comparison of the DeepONet model with FCN and CNN for the cases of $\rm R^2$ takes highest or lowest}
\label{tab:ode_best_worst}
\begin{adjustbox}{width=\textwidth}
\begin{tabular}{@{}lllllll@{}}
\toprule
Test ID                     & Models   & $\rm R^2$         & MSE       & RMSE      & MAE       & RMSE/MAE  \\ \midrule
\multirow{3}{*}{122 (Highest)} & DeepONet & $ 1.000  $&$ 2.000\times 10^{-6} $&$ 1.291\times 10^{-3} $&$ 1.002\times 10^{-3} $&$ 1.289$ \\
                            & FCN      & $ 9.996\times 10^{-1}  $&$ 4.848\times 10^{-5} $&$ 6.962\times 10^{-3} $&$ 5.300\times 10^{-3} $&$ 1.314$ \\
                            & CNN      & $ 9.969\times 10^{-1}  $&$ 4.107\times 10^{-4} $&$ 2.027\times 10^{-2} $&$ 1.194\times 10^{-2} $&$ 1.395$ \\ \midrule
\multirow{3}{*}{72 (Lowest)} & DeepONet & $ 5.316\times 10^{-1}  $&$ 4.200\times 10^{-5} $&$ 6.462\times 10^{-3} $&$ 4.531\times 10^{-3} $&$ 1.426$ \\
                            & FCN      & $ 9.484\times 10^{-1}  $&$ 2.003\times 10^{-6} $&$ 1.415\times 10^{-3} $&$ 9.501\times 10^{-4} $&$ 1.490$ \\
                            & CNN      & $ -7.758 $&$ 3.398\times 10^{-4} $&$ 1.843\times 10^{-2} $&$ 1.769\times 10^{-2} $&$ 1.042$ \\ \bottomrule
\end{tabular}
\end{adjustbox}
\end{table}

\subsection{Diffusion System}
\label{sec:dr}
The second problem is a diffusion system, commonly encountered in various engineering fields, including heat transfer, chemical reactions, and neutron diffusion. The focus is on a time-dependent 1-dimensional diffusion equation given by:

\begin{equation}
\frac{\partial s}{\partial t} = D \frac{\partial ^{2} s}{\partial x^{2}} + ks^{2} + u(x)
\end{equation}

In this equation, with $D = 0.01$ representing the diffusion coefficient and $k = 0.05$ denoting the reaction rate, the domain is defined as $x \in (0,1)$ and $t \in (0,1]$. The input function $u(x)$ is considered as the source term. To generate diverse input functions, we utilize a 1-dimensional Gaussian Random Field (GRF) to create 10,000 patterns of $u(x)$ at 100 fixed sensor positions.

\begin{figure}[!htbp]
    \centering
    \includegraphics[width=13cm]{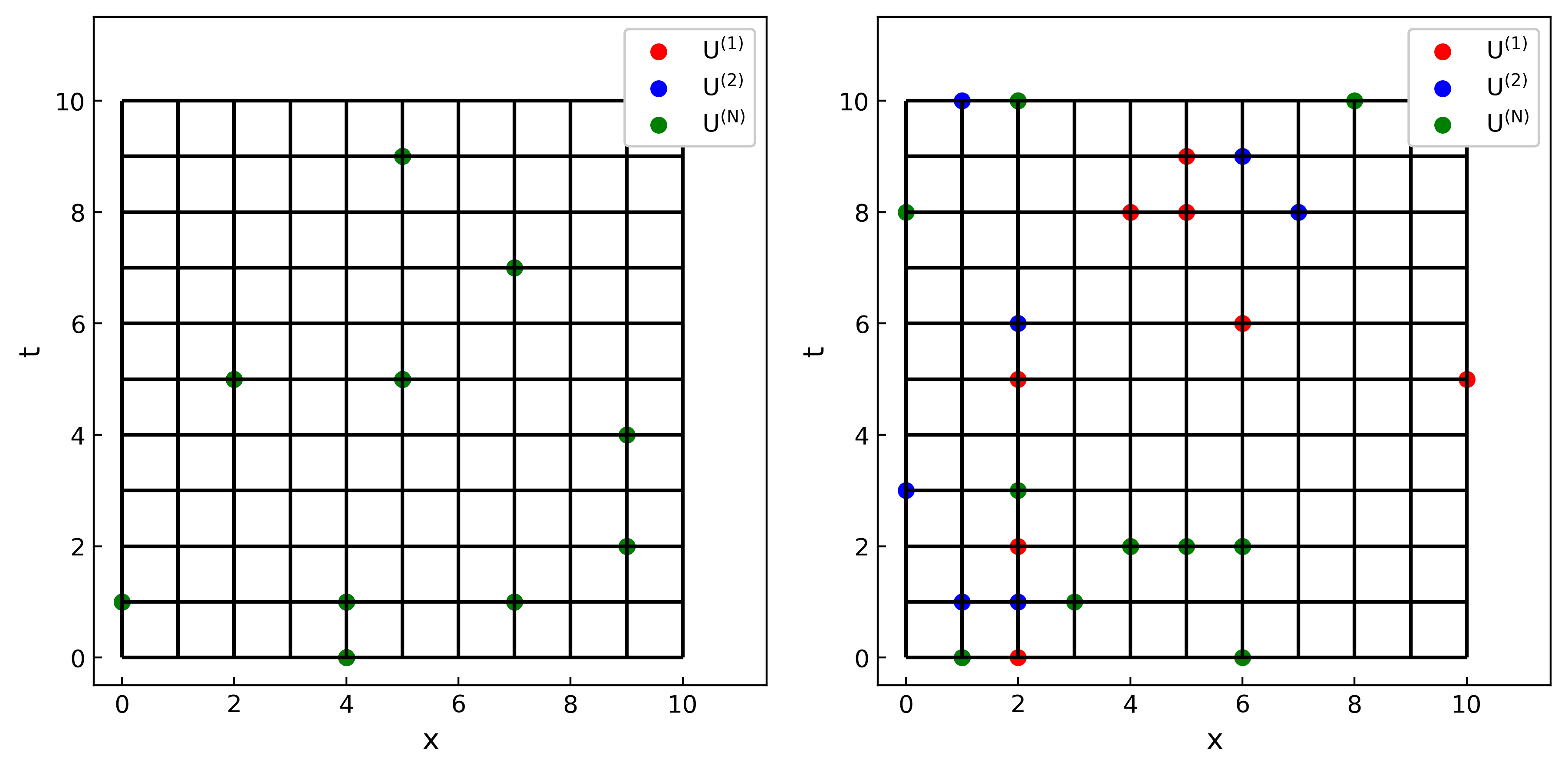}
    \caption{Concept of random selected points $P(x,t)$ (dots) on a $10\times10$ grid for the input variable space of $x \in (0,10)$ and $t \in (0,10)$. The numerical solutions $u(x,t)$ are sampled at the points. The left pane represents a dataset called "aligned dataset," and  the right pane is "unaligned dataset" \cite{DeepONet_Lu}.}
    \label{fig:sample_grid}
\end{figure}

Numerical solutions $s(x,t)$ are obtained for each input function by applying the finite difference method (FDM) on a $100 \times 100$ grid. The process involves randomly sampling 100 points from the grid $(x,y)$, and this sampling procedure is repeated 100 times for each input function. Consequently, a set of numerical solutions is generated. It is worth noting that the random grid points used for sampling differ for each input function $u(x)$. This type of dataset is commonly referred to as an "unaligned dataset" \cite{lu2021deepxde}, and the left panel of Figure \ref{fig:sample_grid} provides a visual representation of this concept. The total size of the training data is determined by the product of the number of input functions ($10^4$) and the number of samplings ($10^2$), resulting in a training data size of $10^6$.

DeepONet employs fully connected neural networks for branch and trunk networks, with a branch network size of [100, 40, 40]. In this problem, the output function $s$ takes a two-dimensional input $(x,t)$, leading to a layer size adjustment [2, 40, 40]. During the training process, the Adam optimization algorithm is utilized for optimization. The mean squared error (MSE) is employed as the evaluation metric to assess the model's performance. The model undergoes training for 10,000 iterations, with a learning rate set to 0.001.

\subsubsection{Results of Dissusion System}
\begin{figure}[!htbp]
    \centering
    \includegraphics[width=13cm]{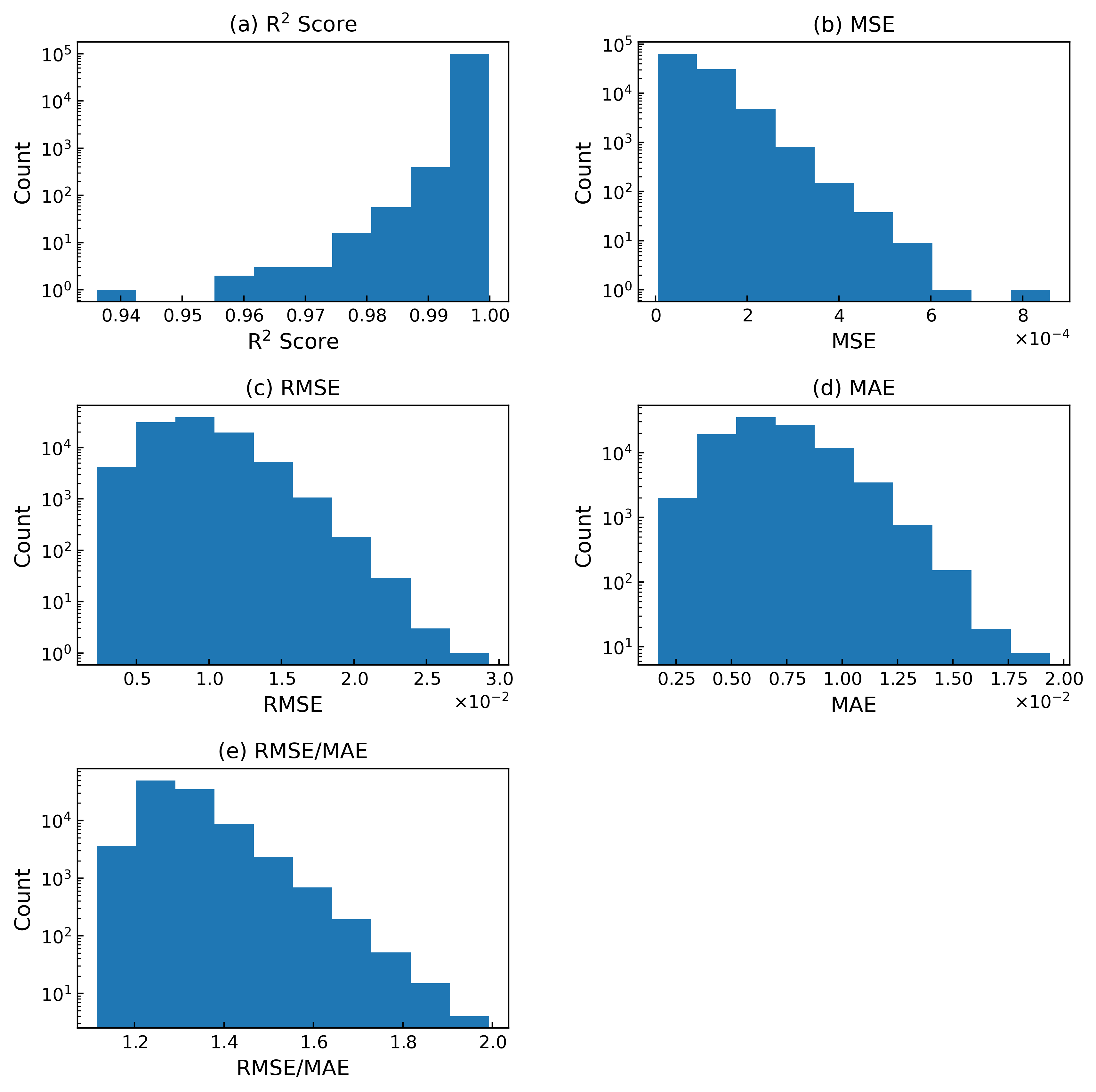}
    \caption{Distribution of each metric over the test
data for the diffusion system (Section \ref{sec:dr}).}
    \label{fig:diffusion_eval}
\end{figure}

Table \ref{tab:diffusion_all} summarizes the performance metrics for the DeepONet model on the test data. The model shows remarkable convergence in $\rm R^2$, with an average score of approximately 0.999, indicating strong agreement between predicted and actual values. Figure \ref{fig:diffusion_eval} shows the Distribution of the performance metrics, and even the lowest $\rm R^2$ scores depicted in Figure \ref{fig:diffusion_eval} (a) are above 0.9, demonstrating the robustness of the model's performance for this problem setup. Moreover, the values of MSE, RMSE, and MAE are significantly low, ranging from $10^{-4}$ to $10^{-2}$, as observed in Figures \ref{fig:diffusion_eval} (b), (c), and (d), indicating high accuracy and precision of the DeepONet model in capturing the diffusion behavior. This is consistent with the performance achieved in the ODE test case described in Section \ref{sec:ode}.

The mean ratio of RMSE to MAE, calculated as 1.300 and listed in Table \ref{tab:diffusion_all} and shown in Figure \ref{fig:diffusion_eval} (e), suggests that some data may be significantly different from the predictions. However, the high $\rm R^2$ score ensures the overall validity of the model.

\begin{figure}[!htbp]
    \centering
    \includegraphics[width=13cm]{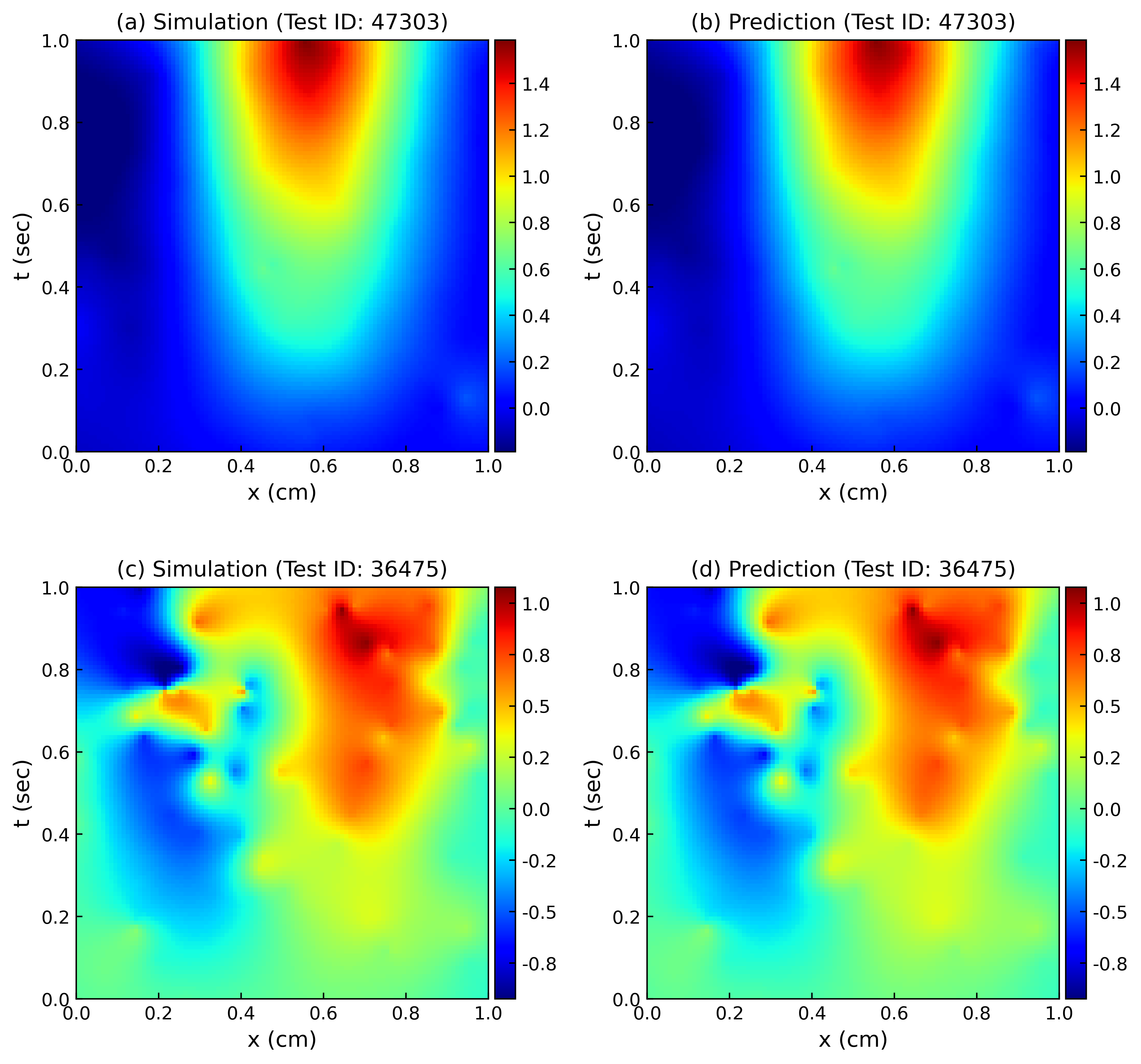}
    \caption{Comparison between simulations and DeepONet predictions of the diffusion system. The test cases with IDs 47303 and 36475 correspond to the highest and lowest $\rm R^2$ scores achieved by DeepONet, respectively. The right figures show prediction results obtained by the operator $G: u(x) \mapsto s(x,t)$.}
    \label{fig:diffusion_test}
\end{figure}

A comprehensive evaluation of the models' performance was conducted by comparing the highest and lowest prediction accuracies with respect to FCN and CNN. The results are summarized in Table \ref{tab:diffusion_best_worst}, providing valuable insights into the differential performance of DeepONet compared to traditional methods. Notably, for Test ID 47303, which achieves the highest $\rm R^2$ score, the predictions of DeepONet closely align with the results obtained from FCN and CNN. Conversely, in the case of Test ID 36475, where FCN and CNN fail to reproduce the test data, DeepONet can accurately predict the desired outcomes accurately. This is evident from the agreement between the DeepONet predictions and the simulation results in Figures \ref{fig:diffusion_test} (c) and (d) for Test ID 36475.

\begin{table}[!htbp]
\centering
\caption{Overall performance metrics of the DeepONet model for the diffusion system}
\label{tab:diffusion_all}
\begin{adjustbox}{width=\textwidth}
\begin{tabular}{@{}llllll@{}}
\toprule
Statistics & $\rm R^2$        & MSE       & RMSE      & MAE       & RMSE/MAE  \\ \midrule
Mean       & $9.990 \times 10^{-1}$ & $8.600 \times 10^{-5}$ & $8.889 \times 10^{-3}$ & $6.838 \times 10^{-3}$ & $1.300$ \\
Std        & $1.087 \times 10^{-3}$ & $5.200 \times 10^{-5}$ & $2.612 \times 10^{-3}$ & $1.974 \times 10^{-3}$ & $7.326 \times 10^{-2}$ \\
Min        & $9.361 \times 10^{-1}$ & $5.000 \times 10^{-6}$ & $2.272 \times 10^{-3}$ & $1.678 \times 10^{-3}$ & $1.116$ \\
Max        & $9.999 \times 10^{-1}$ & $8.600 \times 10^{-4}$ & $2.932 \times 10^{-2}$ & $1.938 \times 10^{-2}$ & $1.993$ \\ \bottomrule
\end{tabular}
\end{adjustbox}
\end{table}

\begin{table}[!htbp]
\centering
\caption{Comparison of the DeepONet model with FCN and CNN for the cases of $\rm R^2$ takes highest or lowest}
\label{tab:diffusion_best_worst}
\begin{adjustbox}{width=\textwidth}
\begin{tabular}{@{}lllllll@{}}
\toprule
Test ID & Models & $\rm R^2$         & MSE       & RMSE      & MAE       & RMSE/MAE  \\ \midrule
\multirow{3}{*}{47303 ($\rm R^2$ Highest)}  & DeepONet & $9.999 \times 10^{-1}$ & $1.345 \times 10^{-5}$ & $3.668 \times 10^{-3}$ & $2.909 \times 10^{-3}$ & 1.261 \\
        & FCN    & $9.934 \times 10^{-1}$  & $1.209 \times 10^{-3}$ & $3.477 \times 10^{-2}$ & $1.853 \times 10^{-2}$ & 1.876 \\
        & CNN    & $9.968 \times 10^{-1}$  & $7.178 \times 10^{-4}$ & $2.679 \times 10^{-2}$ & $1.562 \times 10^{-2}$ & 1.716 \\ \midrule
\multirow{3}{*}{36475 ($\rm R^{2}$ Lowest)} & DeepONet & $9.361 \times 10^{-1}$ & $1.376 \times 10^{-5}$ & $3.710 \times 10^{-3}$ & $2.740 \times 10^{-3}$ & 1.354 \\
        & FCN    & $9.844 \times 10^{-1}$ & $6.517 \times 10^{-4}$ & $2.553 \times 10^{-2}$ & $1.301 \times 10^{-1}$ & 1.950 \\
        & CNN    & $9.941 \times 10^{-1}$ & $2.468 \times 10^{-4}$ & $1.571 \times 10^{-2}$ & $1.016 \times 10^{-2}$ & 1.546 \\ \bottomrule
\end{tabular}
\end{adjustbox}
\end{table}

\subsection{Convection-Diffusion Reaction System}
\label{sec:burger}
This test case considers a conventional diffusion scenario encompassing fluid mechanics, gas dynamics, and nonlinear acoustics. It finds application in the analysis of general industrial products such as small engines, pumps, as well as large turbines used in power plants and aircraft. As an illustrative example, a one-dimensional Burgers' equation equation, which corresponds to the neglect of the pressure term in the Navier-Stokes equation, is selected. The viscous Burgers' equation, describing the problem for a given field $s(x,t)$, is given by:

\begin{equation}
    \frac{\partial s}{\partial t} +  s \frac{\partial s}{\partial x} = \nu \frac{\partial^{2} s}{\partial x^{2}},\,\,\,\,\, x \in (0, 10), t \in (0, 10)
    \label{eq:burger}
\end{equation}

Here, the kinematic viscosity is denoted as $\nu = 0.01$, while $x$ represents the spatial coordinate, and $t$ represents the temporal coordinate. This problem aims to learn the operator $G$, which maps the initial condition $s(x,0)$ to the output function $s(x,t)$. To solve Equation \ref{eq:burger}, the initial condition is specified as $s(x,0) = u(x)$, where $u(x)$ denotes a one-dimensional Gaussian Random Field (GRF).

For the generation of the training dataset, the solution to Equation \ref{eq:burger} is obtained using the Fast Fourier transform (FFT) pseudo-spectral method on a $100 \times 100$ grid. The random locations are fixed for each $u(x)$. This type of dataset is classified as an "aligned dataset" \cite{lu2021deepxde}, as depicted in the left pane of Figure \ref{fig:sample_grid}. The total size of the training dataset amounts to 150, while the test dataset comprises 1,000 instances.

The architecture of ONets is defined by the following: both branch and trunk networks are set to fully connected neural networks. The branch net size is [100, 40, 40]. Like the diffusion system problem in Section \ref{sec:dr}, this problem has two-dimensional input $(x,t)$. Therefore, the layer size of trunk networks is modified as [2, 40, 40]. During the training process, the Adam optimization algorithm is utilized for optimization. The mean L2 relative error is employed as the evaluation metric to assess the model's performance. The model undergoes training for a total of 50,000 iterations, with a learning rate set to 0.001.

\subsubsection{Results of Convection-Diffusion Reaction System}
\begin{figure}[!htbp]
    \centering
    \includegraphics[width=13cm]{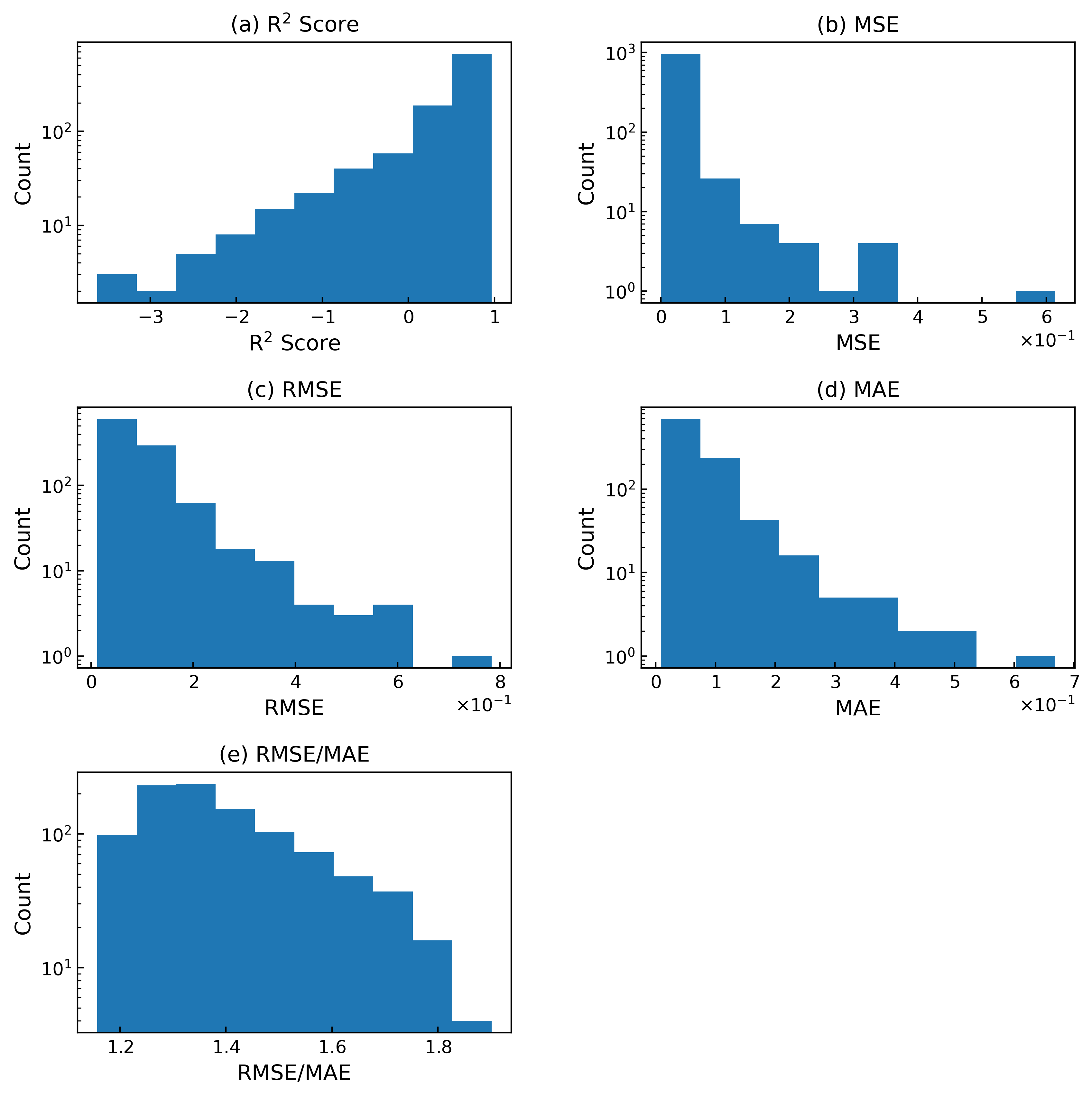}
    \caption{Distribution of each metric over the test
data for the Burgers' equation equation (Section \ref{sec:burger}).}
    \label{fig:burger_eval}
\end{figure}

The summary metrics of the test data for the DeepONet model are presented in Table \ref{tab:burger_all}. The mean $\rm R^2$ score obtained in this study is approximately $0.437$, indicating a relatively low model accuracy. The distributions of metrics are visually represented in Figure \ref{fig:burger_eval}. In particular, Figure \ref{fig:burger_eval} (a) reveals that a significant number of test cases result in negative $\rm R^2$ scores, indicating that the model fails to make accurate predictions. Additionally, the values of MSE, RMSE, and MAE shown in Figures \ref{fig:burger_eval} (b), (c), and (d), respectively, are relatively large at the order of $10^{-1}$, indicating inadequate overall accuracy of the model.

\begin{figure}[!htbp]
    \centering
    \includegraphics[width=13cm]{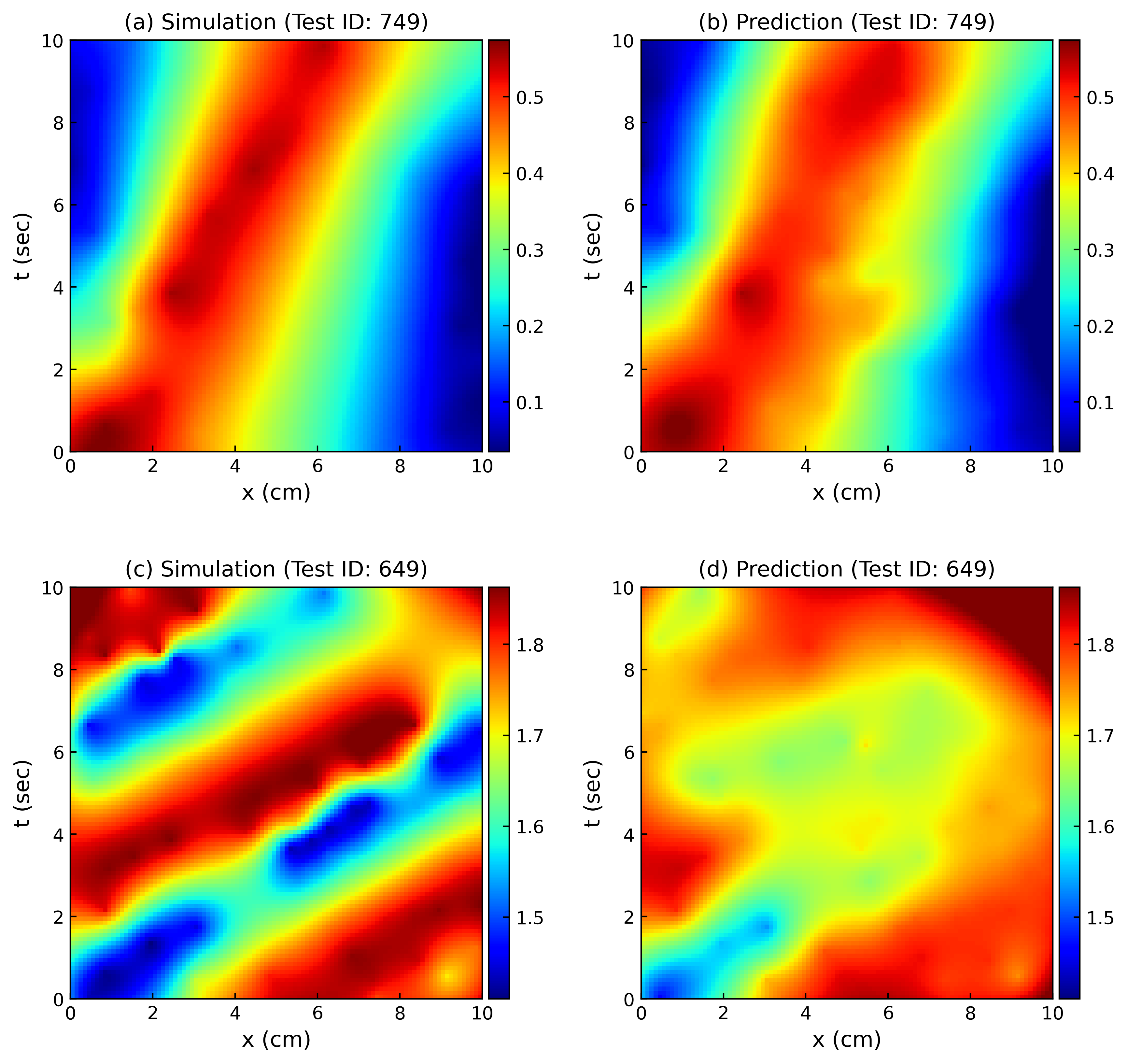}
    \caption{Comparison between simulations and DeepONet predictions of the Burgers' equation equation. The test cases with IDs 749 and 649 correspond to the highest and lowest $\rm R^2$ scores achieved by DeepONet, respectively. The right figures show prediction results obtained by the operator $G: u(x) \mapsto s(x,t)$.}
    \label{fig:burger_test}
\end{figure}

To provide further insights into the performance of DeepONet compared to traditional methods, a comparison of the highest and lowest prediction accuracies with respect to FCN and CNN was conducted. The results are summarized in Table \ref{tab:burger_best_worst}. For Test ID 749, which achieved the highest $\rm R^2$ score of $0.966$, the predictions of DeepONet closely align with the results obtained from FCN and CNN. However, in the case of Test ID 649, which achieved the lowest $\rm R^2$ score of $-3.63$, DeepONet fails to reproduce the test data even when FCN and CNN succeed. These differences in performance are visually demonstrated in Figure \ref{fig:burger_test}, where a comparison between simulations and the model predictions for these test cases is shown. In Figure \ref{fig:burger_test} (a) and (b), it can be observed that DeepONet's prediction for Test ID 749 is not in perfect agreement with the simulation, but the overall trend appears to be reproduced. On the other hand, Figure \ref{fig:burger_test} (c) and (d) clearly indicate that for Test ID 649, DeepONet returns predictions that are quite different from the simulation result.

\begin{table}[!htbp]
\centering
\caption{Overall performance metrics of the DeepONet model for the Burgers' equation equation}
\label{tab:burger_all}
\begin{adjustbox}{width=\textwidth}
\begin{tabular}{@{}llllll@{}}
\toprule
Statistics & $\rm R^2$        & MSE       & RMSE      & MAE       & RMSE/MAE  \\ \midrule
Mean       & $4.365 \times 10^{-1}$ & $1.532 \times 10^{-2}$ & $9.736 \times 10^{-2}$ & $7.079 \times 10^{-2}$ & $1.391$ \\
Std        & $6.461 \times 10^{-1}$ & $3.769 \times 10^{-2}$ & $7.649 \times 10^{-2}$ & $5.947 \times 10^{-2}$ & $1.445 \times 10^{-1}$ \\
Min        & $-3.615$ & $1.510 \times 10^{-4}$ & $1.229 \times 10^{-2}$ & $8.8880 \times 10^{-3}$ & $1.157$ \\
Max        & $9.661 \times 10^{-1}$ & $6.141 \times 10^{-1}$ & $7.837 \times 10^{-1}$ & $6.686 \times 10^{-1}$ & $1.901$ \\ \bottomrule
\end{tabular}
\end{adjustbox}
\end{table}

\begin{table}[!htbp]
\centering
\caption{Comparison of the DeepONet model with FCN and CNN for the cases of $\rm R^2$ takes highest or lowest}
\label{tab:burger_best_worst}
\begin{adjustbox}{width=\textwidth}
\begin{tabular}{@{}lllllll@{}}
\toprule
Test ID & Models & $\rm R^2$         & MSE       & RMSE      & MAE       & RMSE/MAE  \\ \midrule
\multirow{3}{*}{749 ($\rm R^2$ Highest)}  & DeepONet & $9.661 \times 10^{-1}$ & $9.000 \times 10^{-4}$ & $3.000 \times 10^{-2}$ & $2.342 \times 10^{-2}$ & 1.281 \\
        & FCN    & $9.989 \times 10^{-1}$  & $2.837 \times 10^{-5}$ & $5.326 \times 10^{-3}$ & $3.673 \times 10^{-3}$ & 1.450 \\
        & CNN    & $9.989 \times 10^{-1}$  & $2.852 \times 10^{-5}$ & $5.341 \times 10^{-3}$ & $3.864 \times 10^{-3}$ & 1.382 \\ \midrule
\multirow{3}{*}{649 ($\rm R^{2}$ Lowest)} & DeepONet & $-3.615$ & $2.003 \times 10^{-2}$ & $1.415 \times 10^{-1}$ & $1.112 \times 10^{-1}$ & 1.272 \\
        & FCN    & $9.268 \times 10^{-1}$ & $1.583 \times 10^{-3}$ & $3.978 \times 10^{-2}$ & $2.359 \times 10^{-2}$ & 1.686 \\
        & CNN    & $9.830 \times 10^{-1}$ & $3.669 \times 10^{-4}$ & $1.916 \times 10^{-2}$ & $1.316 \times 10^{-2}$ & 1.456 \\ \bottomrule
\end{tabular}
\end{adjustbox}
\end{table}

\section{Discussion}

Based on the findings in Sections \ref{sec:dr} and \ref{sec:burger}, it’s evident that the accuracy of the model for the convection-diffusion-reaction (Burgers' equation) system is relatively lower. While the lack of training iterations or a smaller neural network size might be suspected initially, experiments in \ref{sec:num_train} and \ref{sec:trunk_net} indicate that modifying these parameters does not significantly impact the model's performance on the test data. This leads to the conclusion that the primary issue likely resides in the training data.

To mitigate this issue, augmenting the training data or enhancing the sampling of points $(x, y)$ for the output functions could be beneficial. However, practical constraints in applying these solutions to complex physical systems must be considered, especially in domains like nuclear systems modeling. The physical space limitations and severe operational conditions can impose restrictions on the number and installation of sensors, making increasing the number of sampling points for improved training data acquisition not always feasible. Therefore, exploring alternative approaches to enhance the model’s performance without solely relying on increased sampling or unaligned datasets is imperative.

Considering potential deviations of the model's predictions from the targets, when examining a fluid water system in a pipe with a water temperature of $T=293 \, K$, the kinematic viscosity of water, $\nu=0.01 \, m^2/s$, is equivalent to the problem setup. Erosion caused by fluid represents a significant concern under these assumptions. The solution of the Burgers' equation, denoted as $s(x,t)$, provides the water speed at specific spatial and temporal coordinates, necessitating the acquisition of the fluid velocity distribution. The erosion risk is evaluated by determining whether the fluid speed exceeds the erosion velocity threshold. Ensuring accurate predictions of fluid velocity distribution is crucial due to the potential impacts of erosion in fluid systems. Inaccuracies can compromise the assessment of erosion risk and lead to the implementation of inadequate safety measures. The erosion velocity, $V_e$, can be calculated using the formula $V_e = \frac{C}{\sqrt{\rho}}$, where $C$ is an empirical constant of 240, and $\rho$ is the fluid density, yielding an erosion velocity of $7.6 \, m/s$.

\begin{figure}[!htbp]
    \centering
    \includegraphics[width=8cm]{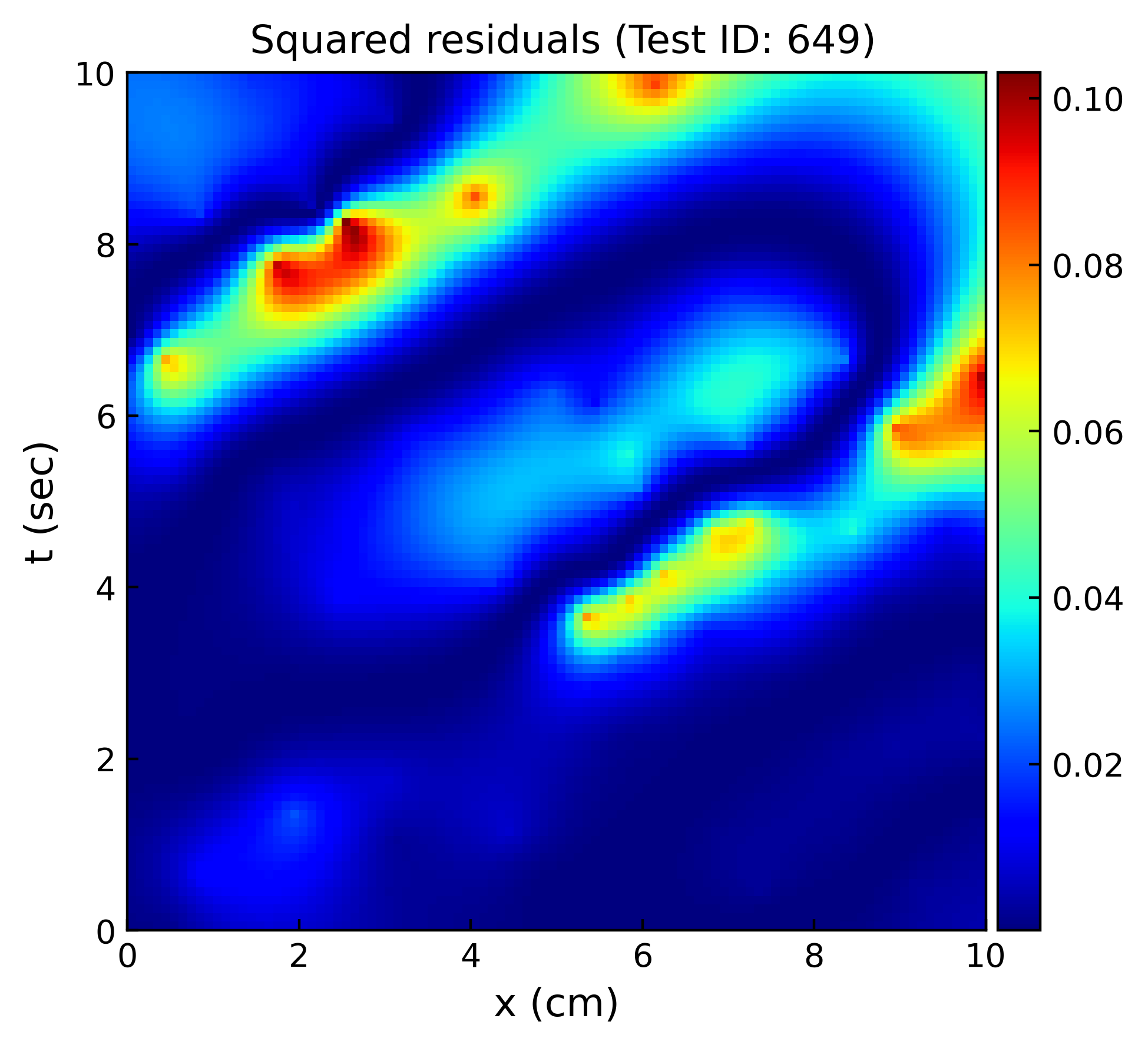}
    \caption{The squared residuals were computed between the simulation and the DeepONet prediction for Test ID 649. The largest value gets at $(x,t)=(2.53, 8.28)$, where the true fluid velocity is $1.45$ $\rm m/s$.}
    \label{fig:burger_squared_residuals}
\end{figure}

Figure \ref{fig:burger_test} (c) illustrates that the fluid velocity remains below the threshold throughout the simulation. However, the most egregious prediction error by DeepONet is depicted in Figure \ref{fig:burger_test} (d). The squared residuals in Figure \ref{fig:burger_squared_residuals} accentuate the discrepancy between the simulation and DeepONet predictions. The most significant difference occurs at $(x,t)=(2.53, 8.28)$, with the actual fluid velocity at $1.45 \, m/s$, and the prediction ratio is 1.22. While this doesn’t exceed the threshold, the predicted erosion risk is elevated by a factor of 1.22.

Given the paramount importance of accuracy in erosion risk assessment, ensuring the model’s reliability and capturing fluid behavior is crucial, especially for systems where accidents can have a profound societal impact, such as power plants and aircraft. In such cases, rigorous validation and verification of the model are indispensable.

\section{Conclusions}

Deep Operator Networks (DeepONets) offer a promising approach for learning solution operators to partial differential equations (PDEs) from data. This study evaluates DeepONets on three test cases: a system of ODEs, a general diffusion system, and the convection-diffusion equation. Accurate predictions are achieved for the ODEs and diffusion cases, with $\rm R^{2}$ scores above 0.96 over the observed domain. However, the convection-diffusion case requires further refinements to the DeepONet algorithm. Nonetheless, the results showcase DeepONets' feasibility as a prediction module for digital twin applications. Going forward, verification, validation, and uncertainty quantification remain critical to ensure robust and reliable surrogate models. This study motivates the integration of neural operators with Bayesian and non-Bayesian filters for real-time digital twin updates. Key future work will focus on accelerating DeepONet predictions for real-time usage and rigorous uncertainty characterization. This study represents an important advance toward generalizable surrogates for spatiotemporal systems. The results confirm the promise of neural operators for digital twins while highlighting essential areas for continued research and maturation.

\section*{Acknowledgement}
The computational part of this work was supported in part by the National Science Foundation (NSF) under Grant No. OAC-1919789.

\newpage
\appendix
\section{Effect of Number of Training Iterations}
\label{sec:num_train}
The sensitivity of the number of training iterations to the DeepONet model for the convection-diffusion-reaction system described in Section \ref{sec:burger} was investigated. Fully connected neural networks were used for the branch and trunk networks, with sizes [100, 40, 40] and [2, 40, 40], respectively. The model's performance was evaluated using mean L2 relative error, and the Adam optimization algorithm was employed. The model underwent training for 100,000 iterations with a learning rate of 0.001.

\begin{figure}[!htbp]
    \centering
    \includegraphics[width=8cm]{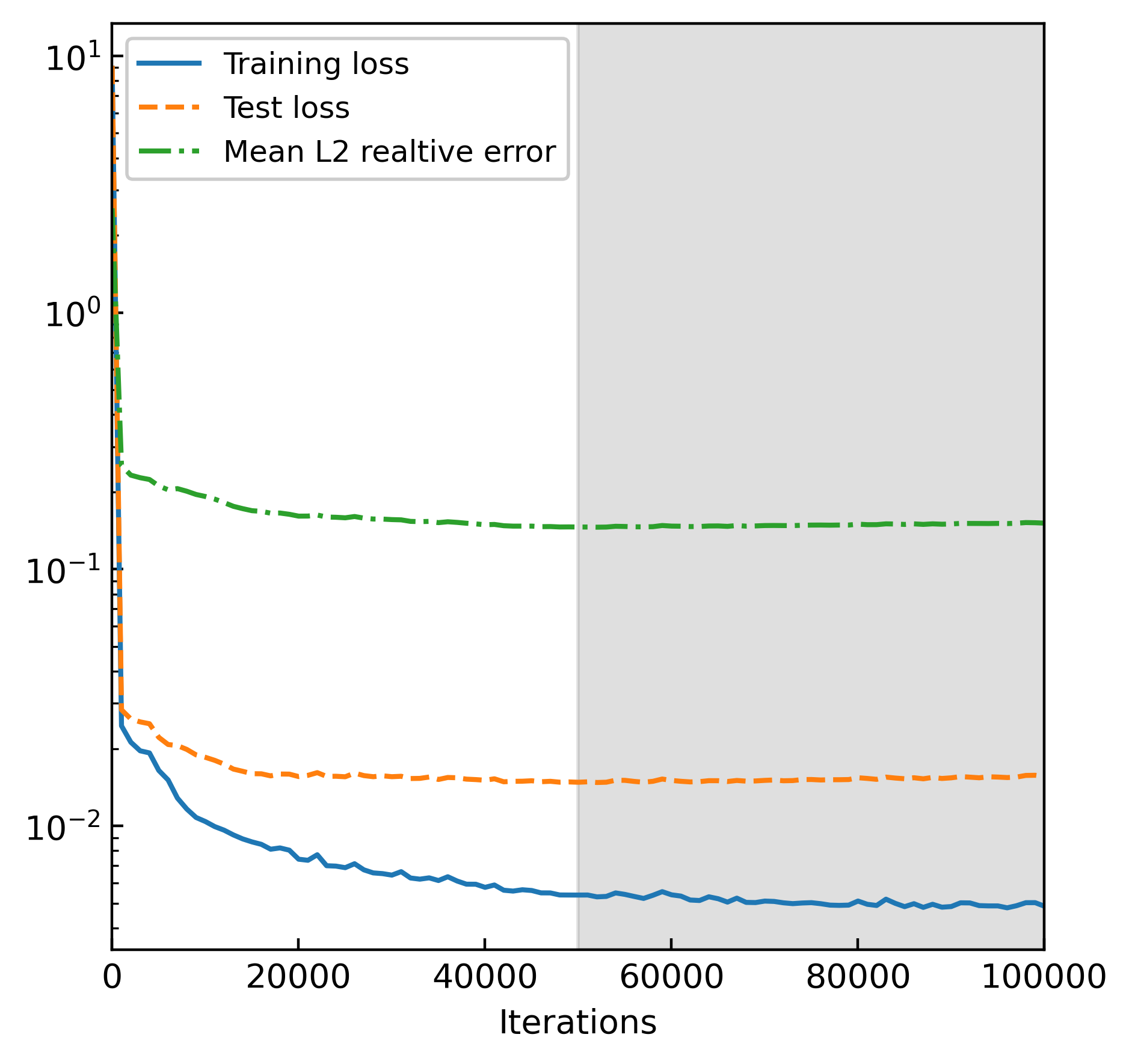}
    \caption{Correlations between the number of iterations and the training loss, test loss, and test metric. The shaded region represents corresponds to over 50,000 iterations.}
    \label{fig:burger_train_hist}
\end{figure}

Correlations between the number of iterations and the training loss, test loss, and test metric were illustrated in Figure \ref{fig:burger_train_hist}. It was observed that the training loss continued to decrease with iterations, while the values of test loss and test metrics reached a plateau after 50,000 iterations.

\newpage
\section{Effect of Trunk Network's Architecture}
\label{sec:trunk_net}
The effect of different trunk network architectures on the convection-diffusion reaction system (Section \ref{sec:burger}) was investigated. The branch network size was fixed at [100, 40, 40], while the hidden layer of the trunk network varied from [2, 20, 40] to [2, 80, 40] with increments of 20 neurons. Model metrics were computed using mean L2 relative error and training employed the Adam optimization algorithm. The models underwent 50,000 iterations with a learning rate of 0.001. However, the results in Table \ref{tab:trunk_net} indicate that modifying the number of neurons in the trunk network's hidden layer did not significantly improve the model metrics.

\begin{table}[!htbp]
\centering
\caption{Impact of the branch network size. The number of neurons in the hidden layer is varied.}
\label{tab:trunk_net}
\begin{adjustbox}{width=\textwidth}
\begin{tabular}{@{}clllll@{}}
\toprule
Neurons & $\rm R^2$        & MSE       & RMSE      & MAE       & RMSE/MAE  \\ \midrule
20      & $4.018\times 10^{-1}$ & $2.154\times 10^{-2}$ & $1.022\times 10^{-1}$ & $7.335\times 10^{-2}$ & 1.418 \\
40      & $4.365\times 10^{-1}$ & $1.532\times 10^{-2}$ & $9.736\times 10^{-2}$ & $7.079\times 10^{-2}$ & 1.391 \\
60      & $4.000\times 10^{-1}$ & $2.195\times 10^{-2}$ & $1.119\times 10^{-1}$ & $8.294\times 10^{-2}$ & 1.371 \\
80      & $3.889\times 10^{-1}$ & $1.810\times 10^{-2}$ & $1.019\times 10^{-1}$ & $7.456\times 10^{-2}$ & 1.379 \\ \bottomrule
\end{tabular}
\end{adjustbox}
\end{table}

\newpage
\section{Comparisons between DeepONet and FCN/CNN}
\label{sec:comparison}
A comparative analysis was conducted between the DeepONet and conventional neural network architectures, Fully-Connected Networks (FCNs) and Convolutional Neural Networks (CNNs). All neural network models were developed using the PyTorch 2.0 deep learning library to ensure standardized implementation and leverage optimized routines. This provided a unified framework to impartially assess the performance of DeepONet against established function approximators on the test problems.

\subsection{Data Preparation}
The following is an explanation of how the data for training and testing FCNs and CNNs was prepared using the example of the diffusion system.

For validation of DeepONet model, 100,000 unseen input functions were generated as the test data. The input variable was defined as the point $P_{i} = (x_{i},t_{i})$ representing the spatial coordinate $x$ and time $t$ in the diffusion system domain. The DeepONet model's $\rm R^2$ performance was computed for each test function using the test data. The input functions yielding the maximal and minimal $\rm R^2$ were identified (Test IDs 47303 and 36475). Given an input function $u$ (e.g., corresponding to Test ID), the DeepONet operator $G(u)$ maps the input variable $P = (x,t)$ to the output $s(P)$. This allows formulating the problem conventionally as approximating $s$ from the input vector $P$ using NNs.

\begin{figure}[!htbp]
    \centering
    \includegraphics[width=10cm]{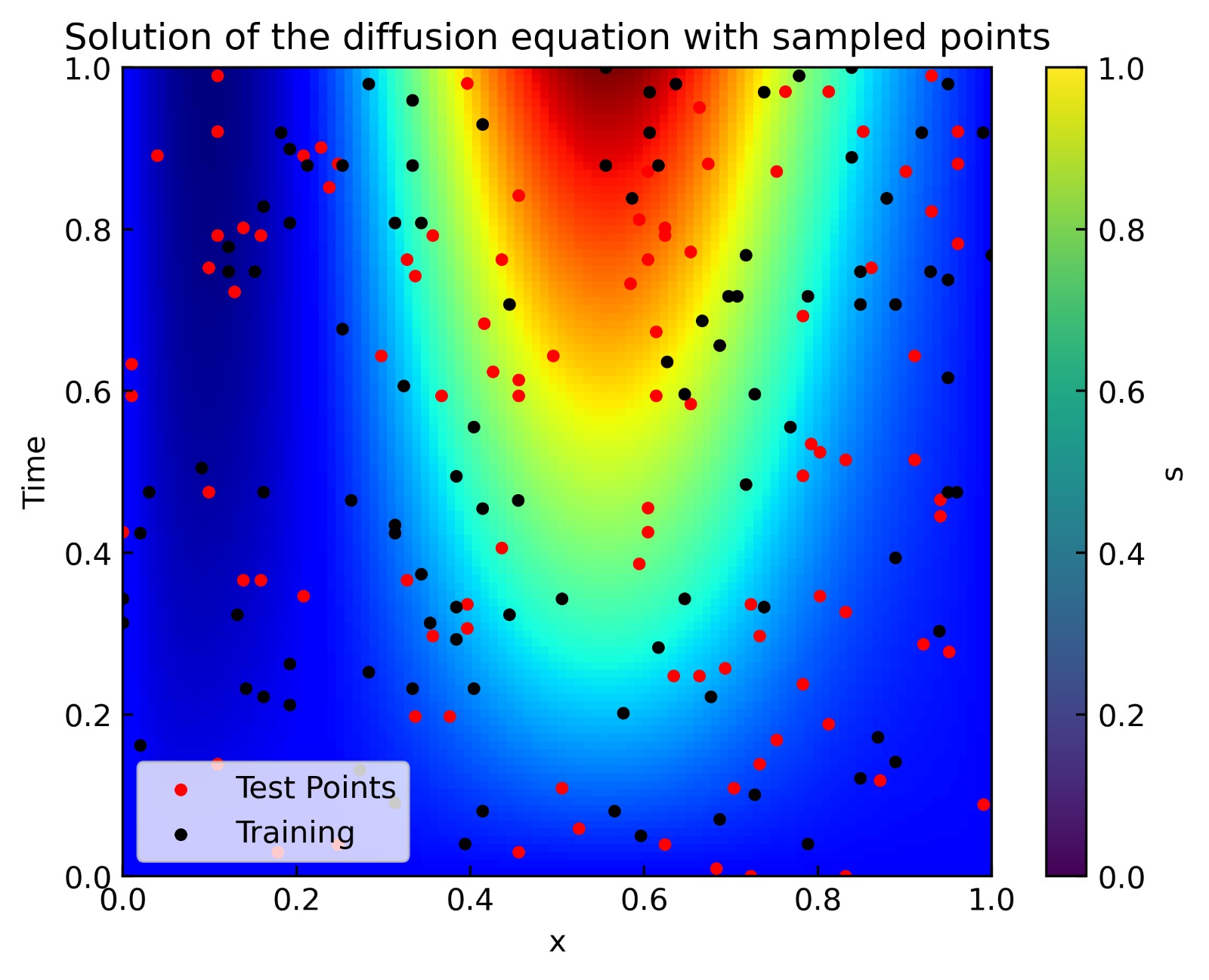}
    \caption{Sampling of training and test datasets for FCN and CNN. The red dots represent the sampled data for Testing DeepONet, FCN, and CNN. The black dots are the training data points for FCN and CNN.}
    \label{fig:resampling}
\end{figure}

In this study, the finite difference method obtained reference solutions $s(x,t)$ by solving the diffusion equation on a $100 \times 100$ grid. As shown in Figure \ref{fig:resampling}, 100 points $P_{i}$ and corresponding $s(P_{i})$ were sampled to create the test data. To train the FCNs and CNNs models for a given input function $u$, the $P_{i}-s(P_{i})$ pairs were resampled excluding points used in the DeepONet's test set. 

For each ODE, Diffusion, and Convection Diffusion demonstration, the training data for FCN and CNN were generated by resampling 50, 100, and 100 on the input variables for the DeepONet test.

\subsection{Network Architectures and Training Parameters}
The network architectures and training parameters for the FCN and CNN models used in the comparative studies are described.

\subsubsection{FCN}
The FCN architectures comprised an input layer of $n$ nodes, two hidden layers with 30 rectified linear unit (ReLU) activated nodes, and an output layer of 1 node for regression. The input layer's size varies depending on the test cases; the ODE problem has $n=1$, and Diffusion and Burgers' equation has $n=2$. Mean squared error (MSE) loss was optimized using the Adam algorithm with a learning rate 0.001 during training. Mini-batch gradient descent was implemented for 2000 epochs with a batch size of 10 samples. Parameters of the training process are listed in Table \ref{tab:training_params}.
\\

\begin{adjustbox}{width=\textwidth}
\begin{lstlisting}
FCN(
  (activation): ReLU()
  (linears): ModuleList(
    (0): Linear(in_features=n_inp, out_features=30, bias=True)
    (1-2): 2 x Linear(in_features=30, out_features=30, bias=True)
    (3): Linear(in_features=30, out_features=1, bias=True)
  )
)
\end{lstlisting}
\end{adjustbox}

\subsection{CNN}
The CNN architectures comprised an input layer of 1 nodes, one 1-dimensional convolution layer which generates 30 output nodes, two linear layer with 30 nodes, and an output layer of 1 node for regression. Mean squared error (MSE) loss was optimized using the Adam algorithm with a learning rate of 0.001 during training. Mini-batch gradient descent was implemented for 2000 epochs with a batch size of 20 samples. Parameters of training process are listed in Table \ref{tab:training_params}.
\\

\begin{adjustbox}{width=\textwidth}
\begin{lstlisting}
CNN(
  (activation): Tanh()
  (cnn1): Conv1d(1, 32, kernel_size=(1,), stride=(1,))
  (flat): Flatten(start_dim=1, end_dim=-1)
  (dropout): Dropout(p=0.5, inplace=False)
  (lin1): Linear(in_features=32, out_features=2048, bias=True)
  (lin2): Linear(in_features=2048, out_features=1, bias=True)
)
\end{lstlisting}
\end{adjustbox}

\begin{table}[htbp]
\caption{Parameters of training for FCN and CNN}
\begin{adjustbox}{width=\textwidth}
\label{tab:training_params}
\begin{tabular}{@{}llllll@{}}
\toprule
\multirow{2}{*}{Test Case} & \multicolumn{5}{c}{FCN/CNN}                                    \\ \cmidrule(l){2-6} 
                           & Loss function & Optimizer & Leaning rate & Epochs & Batch size \\ \midrule
(a) ODE                    & MSE           & Adam      & 0.001        & 2000   & 10/20      \\
(b) Diffusion              & MSE           & Adam      & 0.001        & 2000   & 10/20      \\
(c) Burgers' Equation      & MSE           & Adam      & 0.001        & 2000   & 10/20      \\ \bottomrule
\end{tabular}
\end{adjustbox}
\end{table}

\newpage

\section*{Declaration of Generative AI and AI-assisted technologies in the writing process}
During the preparation of this work the author(s) used ChatGPT September 25 Version in order to language editing and refinement. After using this tool/service, the author(s) reviewed and edited the content as needed and take(s) full responsibility for the content of the publication. [\href{https://www.elsevier.com/about/policies/publishing-ethics/the-use-of-ai-and-ai-assisted-writing-technologies-in-scientific-writing}{Elsevier Publishing Ethics}]

\bibliographystyle{unsrtnat}
\bibliography{references} 

\end{document}